\documentclass[fleqn]{SCYE-arxiv}
\usepackage[numbers,compress]{natbib}

\usepackage{longtable}
\usepackage{mydef}

\begin{document}

\ArticleType{Invited Review}
\Year{2020}
\Month{March}
\Vol{xx}
\No{xx}
\BeginPage{1} 
\EndPage{25}

\title{Pre-trained Models for Natural Language Processing: A Survey}

\author{Xipeng Qiu}{xpqiu@fudan.edu.cn}%
\author{Tianxiang Sun}{}
\author{Yige Xu}{}
\author{Yunfan Shao}{}
\author{Ning Dai}{}
\author{Xuanjing Huang}{}


\AuthorMark{QIU XP, et al.}

\AuthorCitation{QIU XP, et al}

\address{School of Computer Science, Fudan University, Shanghai {\rm 200433}, China}
\address{Shanghai Key Laboratory of Intelligent Information Processing, Shanghai {\rm 200433}, China}


\abstract{Recently, the emergence of pre-trained models (PTMs)\footnote{PTMs are also known as pre-trained language models (PLMs). In this survey, we use PTMs for NLP instead of PLMs to avoid confusion with the narrow concept of probabilistic (or statistical) language models.} has brought natural language processing (NLP) to a new era.
In this survey, we provide a comprehensive review of PTMs for NLP. We first briefly introduce language representation learning and its research progress. Then we systematically categorize existing PTMs based on a taxonomy from four different perspectives. Next, we describe how to adapt the knowledge of PTMs to downstream tasks. Finally, we outline some potential directions of PTMs for future research. This survey is purposed to be a hands-on guide for understanding, using, and developing PTMs for various NLP tasks.}


\keywords{Deep Learning, Neural Network, Natural Language Processing, Pre-trained Model, Distributed Representation, Word Embedding, Self-Supervised Learning, Language Modelling}

\maketitle



\begin{multicols}{2}

\section{Introduction}\label{sec:intro}

With the development of deep learning, various neural networks have been widely used to solve  Natural Language Processing (NLP) tasks, such as convolutional neural networks (CNNs)~\cite{kalchbrenner2014convolutional,kim2014convolutional,gehring2017convolutional}, recurrent neural networks (RNNs)~\cite{sutskever2014sequence,liu2016recurrent}, graph-based neural networks (GNNs)~\cite{socher2013recursive,tai2015improved,DBLP:conf/naacl/MarcheggianiBT18} and attention mechanisms~\cite{bahdanau2014neural,vaswani2017transformer}.
One of the advantages of these neural models is their ability to alleviate the \textit{feature engineering} problem. Non-neural NLP methods usually heavily rely on the discrete handcrafted features, while neural methods usually use low-dimensional and dense vectors (aka. \textit{distributed representation}) to implicitly represent the syntactic or semantic features of the language. These representations are learned in specific NLP tasks. Therefore, neural methods make it easy for people to develop various NLP systems.



Despite the success of neural models for NLP tasks, the performance improvement may be less significant compared to the Computer Vision (CV) field. The main reason is that current datasets for most supervised NLP tasks are rather small (except machine translation). Deep neural networks usually have a large number of parameters, which make them overfit on these small training data and do not generalize well in practice. Therefore, the early neural models for many NLP tasks were relatively shallow and usually consisted of only 1$\sim$3 neural layers.


Recently, substantial work has shown that pre-trained models (PTMs), on the large corpus can learn universal language representations, which are beneficial for downstream NLP tasks and can avoid training a new model from scratch. With the development of computational power, the emergence of the deep models (i.e., Transformer~\cite{vaswani2017transformer}), and the constant enhancement of training skills, the architecture of PTMs has been advanced from shallow to deep.
The \textit{first-generation PTMs} aim to learn good word embeddings. Since these models themselves are no longer needed by downstream tasks, they are usually very shallow for computational efficiencies, such as Skip-Gram \cite{mikolov2013word2vec} and GloVe \cite{DBLP:conf/emnlp/PenningtonSM14}. Although these pre-trained embeddings can capture semantic meanings of words, they are context-free and fail to capture higher-level concepts in context, such as polysemous disambiguation, syntactic structures, semantic roles, anaphora. The \textit{second-generation PTMs} focus on learning contextual word embeddings, such as CoVe~\cite{mccan2017learn}, ELMo~\cite{peters2018elmo}, OpenAI GPT~\cite{radford2018improving} and BERT~\cite{devlin2019bert}. These learned encoders are still needed to represent words in context by downstream tasks. Besides, various pre-training tasks are also proposed to learn PTMs for different purposes.

The contributions of this survey can be summarized as follows:
\begin{enumerate}
\item \textit{Comprehensive review.} We provide a comprehensive review of PTMs for NLP, including background knowledge, model architecture, pre-training tasks, various extensions, adaption approaches, and applications.

\item \textit{New taxonomy.} We propose a taxonomy of PTMs for NLP, which categorizes existing PTMs from four different perspectives: 1) representation type, 2) model architecture; 3) type of pre-training task; 4) extensions for specific types of scenarios.

\item \textit{Abundant resources.} We collect abundant resources on PTMs, including
    open-source implementations of PTMs, visualization tools, corpora, and paper lists.

\item \textit{Future directions.} We discuss and analyze the limitations of existing PTMs. Also, we suggest possible future research directions.
\end{enumerate}


The rest of the survey is organized as follows. Section \ref{sec:bg} outlines the background concepts and commonly used notations of PTMs.
Section \ref{sec:overview} gives a brief overview of PTMs and clarifies the categorization of PTMs. Section \ref{sec:extension} provides extensions of PTMs.
Section \ref{sec:adapt} discusses how to transfer the knowledge of PTMs to downstream tasks.
Section \ref{sec:resources} gives the related resources on PTMs.
Section \ref{sec:app}
presents a collection of applications across various NLP tasks.
Section \ref{sec:future} discusses the current challenges and suggests future
directions. Section \ref{sec:conclusion} summarizes the paper.





\section{Background}
\label{sec:bg}

\subsection{Language Representation Learning}

As suggested by \citet{bengio2013representation}, a good representation should express general-purpose priors that are not task-specific but would be likely to be useful
for a learning machine to solve AI-tasks.
When it comes to language, a good representation should capture the implicit linguistic rules and common sense knowledge hiding in text data, such as lexical meanings, syntactic structures, semantic roles, and even pragmatics.

\begin{figure}[H]
  \centering
 \includegraphics[width=0.48\textwidth]{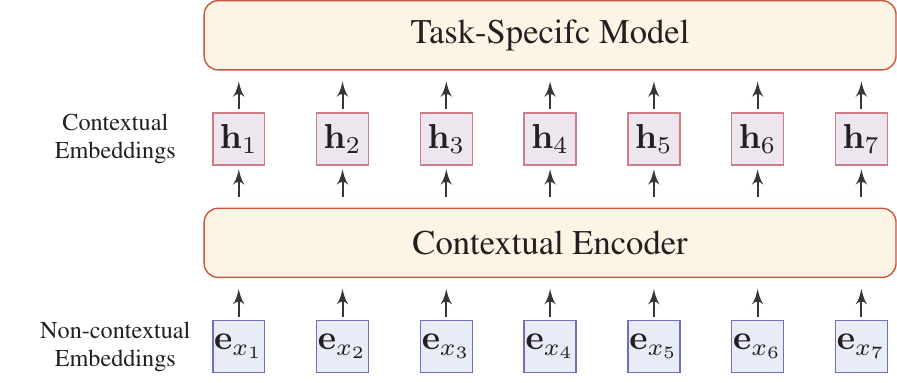}
      \caption{Generic Neural Architecture for NLP}\label{fig:generic}
\end{figure}

The core idea of distributed representation is to describe the meaning of a piece of text by low-dimensional real-valued vectors. And each dimension of the vector has no corresponding sense, while the whole represents a concrete concept.
Figure \ref{fig:generic} illustrates the generic neural architecture for NLP. There are two kinds of word embeddings: non-contextual and contextual embeddings. The difference between them is whether the embedding for a word dynamically changes according to the context it appears in.

\paragraph{Non-contextual Embeddings}

The first step of representing language is to map discrete language symbols into a distributed embedding space. Formally, for each word (or sub-word) $x$ in a vocabulary $\mathcal{V}$, we map it to a vector $\be_{x} \in \mathbb{R}^{D_e}$ with a lookup table $\bE \in \mathbb{R}^{D_e \times |\mathcal{V}|}$, where $D_e$ is a hyper-parameter indicating the dimension of token embeddings. These embeddings are trained on task data along with other model parameters.

There are two main limitations to this kind of embeddings. The first issue is that the embeddings are static. The embedding for a word does is always the same regardless of its context. Therefore, these \textit{non-contextual embeddings} fail to model polysemous words.
The second issue is the out-of-vocabulary problem. To tackle this problem, character-level word representations or sub-word representations are widely used in many NLP tasks, such as CharCNN \cite{kim2016character}, FastText~\cite{DBLP:journals/tacl/BojanowskiGJM17} and Byte-Pair Encoding (BPE) \cite{DBLP:conf/acl/SennrichHB16a}.


\begin{figure*}[t]
  \centering
\subfloat[Convolutional Model]{\label{fig:cnn}
  \includegraphics[width=0.29\textwidth]{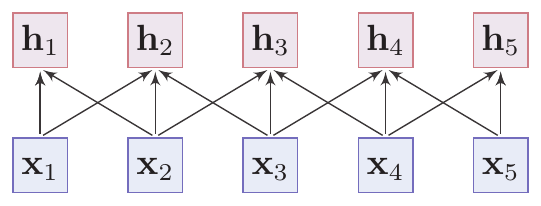}
}
\hspace{2em}
\subfloat[Recurrent Model]{\label{fig:selfatt-rnn}
  \includegraphics[width=0.29\textwidth]{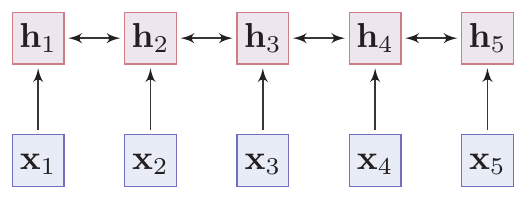}
}
\hspace{2em}
\subfloat[Fully-Connected Self-Attention Model]{\label{fig:selfatt-rnn}
  \includegraphics[width=0.29\textwidth]{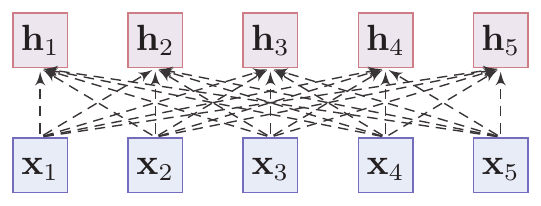}
}
  \caption{Neural Contextual Encoders}
  \label{fig:neural-encoders}
\end{figure*}

\paragraph{Contextual Embeddings}

To address the issue of polysemous and the context-dependent nature of words, we need
distinguish the semantics of words in different contexts. Given a text $x_1,x_2,\cdots,x_T$ where each token $x_t\in \mathcal{V}$ is a word or sub-word, the contextual representation of $x_t$ depends on the whole text.
\begin{align}
[\bh_1,\bh_2,\cdots,\bh_T] = f_{\mathrm{enc}}(x_1,x_2,\cdots,x_T),
\end{align}
where $f_{\mathrm{enc}}(\cdot)$ is neural encoder, which is described in Section~\ref{sec:neural-encoders}, $\bh_t$ is called \textit{contextual embedding} or \textit{dynamical embedding} of token $x_t$ because of the contextual information included in.

\subsection{Neural Contextual Encoders}
\label{sec:neural-encoders}

Most of the neural contextual encoders can be classified into two categories: sequence models and non-sequence models. Figure \ref{fig:neural-encoders} illustrates three representative architectures.

\subsubsection{Sequence Models}

Sequence models usually capture local context of a word in sequential order.

\paragraph{Convolutional Models}
Convolutional models take the embeddings of words in the input sentence and capture the meaning of a word by aggregating the local information from its neighbors by convolution operations~\cite{kim2014convolutional}.

\paragraph{Recurrent Models}
Recurrent models capture the contextual representations of words with short memory, such as LSTMs \cite{DBLP:journals/neco/HochreiterS97} and GRUs \cite{chung2014empirical}.
In practice, bi-directional LSTMs or GRUs are used to collect information from both sides of a word, but its performance is often affected by the long-term dependency problem.


\subsubsection{Non-Sequence Models}

Non-sequence models learn the contextual representation with a pre-defined tree or graph structure between words, such as the syntactic structure or semantic relation. Some popular non-sequence models include Recursive NN~\cite{socher2013recursive}, TreeLSTM~\cite{tai2015improved,zhu2015long}, and GCN~\cite{kipf2017semi}.

Although the linguistic-aware graph structure can provide useful inductive bias, how to build a good graph structure is also a challenging problem. Besides, the structure depends heavily on expert knowledge or external NLP tools, such as the dependency parser.

\paragraph{Fully-Connected Self-Attention Model}
In practice, a more straightforward way is to use a fully-connected graph to model the relation of every two words and let the model learn the structure by itself. Usually, the connection weights are dynamically computed by the self-attention mechanism, which implicitly indicates the connection between words.
A successful instance of fully-connected self-attention model is the Transformer~\cite{vaswani2017transformer,lin2021survey}, which also needs other supplement modules, such as positional embeddings, layer normalization, residual connections and position-wise feed-forward network (FFN) layers.


\subsubsection{Analysis}

Sequence models learn the contextual representation of the word with locality bias and are hard to capture the long-range interactions between words. Nevertheless, sequence models are usually easy to train and get good results for various NLP tasks.

In contrast, as an instantiated fully-connected self-attention model, the Transformer can directly model the dependency between every two words in a sequence, which is more powerful and suitable to model long range dependency of language.
However, due to its heavy structure and less model bias, the Transformer usually requires a large training corpus and is easy to overfit on small or modestly-sized datasets~\cite{radford2018improving,guo2019star}.

Currently, the Transformer has become the mainstream architecture of PTMs due to its powerful capacity.

\subsection{Why Pre-training?}

With the development of deep learning, the number of model parameters has increased rapidly. The much larger dataset is needed to fully train model parameters and prevent overfitting.
However, building large-scale labeled datasets is a great challenge for most NLP tasks due to the extremely expensive annotation costs, especially for syntax and semantically related tasks.

In contrast, large-scale unlabeled corpora are relatively easy to construct. To leverage the huge unlabeled text data, we can first learn a good representation from them and then use these representations for other tasks. Recent studies have demonstrated significant performance gains on many NLP tasks with the help of the representation extracted from the PTMs on the large unannotated corpora.



The advantages of pre-training can be summarized as follows:
\begin{enumerate}
  \item Pre-training on the huge text corpus can learn universal language representations and help with the downstream tasks.
  \item Pre-training provides a better model initialization, which usually leads to a better generalization performance and speeds up convergence on the target task.
  \item Pre-training can be regarded as a kind of regularization to avoid overfitting on small data~\cite{DBLP:journals/jmlr/ErhanBCMVB10}.
\end{enumerate}


\subsection{A Brief History of PTMs for NLP}

Pre-training has always been an effective strategy to learn the parameters of deep neural networks, which are then fine-tuned on downstream tasks.
As early as 2006, the breakthrough of deep learning came with greedy layer-wise unsupervised pre-training followed by supervised fine-tuning~\cite{hinton2006reducing}.
In CV, it has been in practice to pre-train models on the huge ImageNet corpus, and then fine-tune further on smaller data for different tasks.
This is much better than a random initialization because the model learns general image features, which can then be used in various vision tasks.

In NLP, PTMs on large corpus have also been proved to be beneficial for the downstream NLP tasks, from the shallow word embedding to deep neural models.

\subsubsection{First-Generation PTMs: Pre-trained Word Embeddings}

Representing words as dense vectors has a long history~\cite{hinton1986distributed}. The ``modern'' word embedding is introduced in pioneer work of neural network language model (NNLM)~\cite{bengio2003neural}.
\citet{DBLP:journals/jmlr/CollobertWBKKK11} showed that the pre-trained word embedding on the unlabelled data could significantly improve many NLP tasks. To address the computational complexity, they learned word embeddings with \textit{pairwise ranking} task instead of language modeling.
Their work is the first attempt to obtain generic word embeddings useful for other tasks from unlabeled data.
\citet{mikolov2013word2vec} showed that there is no need for deep neural networks to build good word embeddings. They propose two shallow architectures: Continuous Bag-of-Words (CBOW) and Skip-Gram (SG) models.
Despite their simplicity, they can still learn high-quality word embeddings to capture the latent syntactic and semantic similarities among words.
Word2vec is one of the most popular implementations of these models and makes the pre-trained word embeddings accessible for different tasks in NLP.
Besides, GloVe~\cite{DBLP:conf/emnlp/PenningtonSM14} is also a widely-used model for obtaining pre-trained word embeddings, which are computed by global word-word co-occurrence statistics from a large corpus.

Although pre-trained word embeddings have been shown effective in NLP tasks, they are context-independent and mostly trained by shallow models. When used on a downstream task, the rest of the whole model still needs to be learned from scratch.

During the same time period, many researchers also try to learn embeddings of paragraph, sentence or document, such as paragraph vector~\cite{le2014distributed}, Skip-thought vectors~\cite{kiros2015skip}, Context2Vec~\cite{melamud-etal-2016-context2vec}.
Different from their modern successors, these sentence embedding models try to encode input sentences into a fixed-dimensional vector representation, rather than the contextual representation for each token.

\subsubsection{Second-Generation PTMs: Pre-trained Contextual Encoders}


Since most NLP tasks are beyond word-level, it is natural to pre-train the neural encoders on sentence-level or higher. The output vectors of neural encoders are also called \textit{contextual word embeddings} since they represent the word semantics depending on its context.

\citet{dai2015semi} proposed the first successful instance of PTM for NLP. They initialized LSTMs with a language model (LM) or a sequence autoencoder, and found the pre-training can improve the training and generalization of LSTMs in many text classification tasks.
\citet{liu2016recurrent} pre-trained a shared LSTM encoder with LM and fine-tuned it under the multi-task learning (MTL) framework. They found the pre-training and fine-tuning can further improve the performance of MTL for several text classification tasks. \citet{ramachandran2017unsupervised} found the Seq2Seq models can be significantly improved by unsupervised pre-training. The weights of both encoder and decoder are initialized with pre-trained weights of two language models and then fine-tuned with labeled data.
Besides pre-training the contextual encoder with LM, \citet{mccan2017learn} pre-trained a deep LSTM encoder from an attentional sequence-to-sequence model with machine translation (MT). The context vectors (CoVe) output by the pre-trained encoder can improve the performance of a wide variety of common NLP tasks.

Since these precursor PTMs, the modern PTMs are usually trained with larger scale corpora, more powerful or deeper architectures (e.g., Transformer), and new pre-training tasks.

\citet{peters2018elmo} pre-trained 2-layer LSTM encoder with a bidirectional language model (BiLM), consisting of a forward LM and a backward LM.
The contextual representations output by the pre-trained BiLM, ELMo (Embeddings from Language Models), are shown to bring large improvements on a broad range of NLP tasks.
\citet{akbik2018contextual} captured word meaning with contextual string embeddings pre-trained with character-level LM.
However, these two PTMs are usually used as a feature extractor to produce the contextual word embeddings, which are fed into the main model for downstream tasks. Their parameters are fixed, and the rest parameters of the main model are still trained from scratch.
ULMFiT (Universal Language Model Fine-tuning)~\cite{DBLP:conf/acl/RuderH18} attempted to fine-tune pre-trained LM for text classification (TC) and achieved state-of-the-art results on six widely-used TC datasets. ULMFiT consists of 3 phases: 1) pre-training LM on general-domain data; 2) fine-tuning LM on target data; 3) fine-tuning on the target task.
ULMFiT also investigates some effective fine-tuning strategies, including discriminative fine-tuning, slanted triangular learning rates, and gradual unfreezing.

More recently, the very deep PTMs have shown their powerful ability in learning universal language representations: e.g., OpenAI GPT (Generative Pre-training) \cite{radford2018improving} and BERT (Bidirectional Encoder Representation from Transformer) \cite{devlin2019bert}. Besides LM, an increasing number of self-supervised tasks (see Section~\ref{sec:pre-training-tasks}) is proposed to make the PTMs capturing more knowledge form large scale text corpora.

Since ULMFiT and BERT, fine-tuning has become the mainstream approach to adapt PTMs for the downstream tasks.





\section{Overview of PTMs}
\label{sec:overview}

The major differences between PTMs are the usages of contextual encoders, pre-training tasks, and purposes. We have briefly introduced the architectures of contextual encoders in Section \ref{sec:neural-encoders}. In this section, we focus on the description of pre-training tasks and give a taxonomy of PTMs.

\subsection{Pre-training Tasks}
\label{sec:pre-training-tasks}

The pre-training tasks are crucial for learning the universal representation of language. Usually, these pre-training tasks should be challenging and have substantial training data. In this section, we summarize the pre-training tasks into three categories: supervised learning, unsupervised learning, and self-supervised learning.

\begin{enumerate}
  \item \textit{Supervised learning} (SL) is to learn a function that maps an input to an output based on training data consisting of input-output pairs.
  \item \textit{Unsupervised learning} (UL) is to find some intrinsic knowledge from unlabeled data, such as clusters, densities, latent representations.
  \item \textit{Self-Supervised learning} (SSL) is a blend of supervised learning and unsupervised learning\footnote{Indeed, it is hard to clearly distinguish the unsupervised learning and self-supervised learning. For clarification, we refer ``unsupervised learning'' to the learning
without human-annotated supervised labels. The purpose of ``self-supervised learning'' is to learn the general knowledge from data rather than standard unsupervised objectives, such as density estimation.}.
The learning paradigm of SSL is entirely the same as supervised learning, but the labels of training data are generated automatically.
The key idea of SSL is to predict any part of the input from other parts in some form. For example, the masked language model (MLM) is a self-supervised task that attempts to predict the masked words in a sentence given the rest words.

\end{enumerate}



In CV, many PTMs are trained on large supervised training sets like ImageNet.
However, in NLP, the datasets of most supervised tasks are not large enough to train a good PTM. The only exception is machine translation (MT). A large-scale MT dataset, WMT 2017, consists of more than 7 million sentence pairs.
Besides, MT is one of the most challenging tasks in NLP, and an encoder pre-trained on MT can benefit a variety of downstream NLP tasks.
As a successful PTM, CoVe \cite{mccan2017learn} is an encoder pre-trained on MT task and improves a wide variety of common NLP tasks: sentiment analysis (SST, IMDb), question classification (TREC), entailment (SNLI), and question answering (SQuAD).

In this section, we introduce some widely-used pre-training tasks in existing PTMs. We can regard these tasks as self-supervised learning.
Table \ref{tab:loss-function} also summarizes their  loss functions.

\begin{table*}[t]
 \footnotesize \centering
 \begin{threeparttable}\caption{Loss Functions of Pre-training Tasks}\label{tab:loss-function}
 \tabcolsep 10pt 
 \begin{tabular}{llP{0.4\linewidth}}
 \toprule
   Task & Loss Function & Description\\%
   \midrule %
   LM& $\displaystyle\mathcal{L}_{\textrm{\tiny LM}}=-\sum_{t=1}^{T} \log p(x_t|\bx_{<t})$ & $\bx_{<t} = x_1,x_2,\cdots,x_{t-1}$.\\
   \midrule
   MLM& $\displaystyle \mathcal{L}_{\textrm{\tiny MLM}}=-\sum_{\hat{x}\in m(\bx)} \log p\Big(\hat{x}|\bx_{\setminus m(\bx)}\Big)$ &$m(\bx)$ and $\bx_{\setminus m(\bx)}$ denote the masked words from $\bx$ and the rest words respectively.\\
   \midrule
   Seq2Seq MLM& $\displaystyle\mathcal{L}_{\textrm{\tiny S2SMLM}}=-\sum_{t=i}^j \log p\Big(x_t|\bx_{\setminus \bx_{i:j}},\bx_{i:t-1}\Big)$ & $\bx_{i:j}$ denotes an masked n-gram span from $i$ to $j$ in $\bx$.\\
      \midrule
   PLM & $\displaystyle \mathcal{L}_{\textrm{\tiny PLM}}=-\sum_{t=1}^{T} \log p(z_t|\bz_{<t})$  &$\bz=perm(\bx)$ is a permutation of $\bx$ with random order. \\
   \midrule
   DAE & $\displaystyle \mathcal{L}_{\textrm{\tiny DAE}}=-\sum_{t=1}^{T} \log p(x_t|\hat{\bx},\bx_{<t})$ & $\hat{\bx}$ is  randomly perturbed text from $\bx$.\\
   \midrule
   DIM & $\displaystyle \mathcal{L}_{\textrm{\tiny DIM}}= s(\hat{\bx}_{i:j},\bx_{i:j})-\log \sum_{\tilde{\bx}_{i:j}\in \mathcal{N}}s(\hat{\bx}_{i:j},\tilde{\bx}_{i:j})$ & $\bx_{i:j}$ denotes an n-gram span from $i$ to $j$ in $\bx$, $\hat{\bx}_{i:j}$ denotes a sentence masked at position $i$ to $j$, and $\tilde{\bx}_{i:j}$ denotes a randomly-sampled  negative n-gram from corpus.\\
   \midrule
   NSP/SOP & $\displaystyle \mathcal{L}_{\textrm{\tiny NSP/SOP}}=-\log p(t|\bx,\by)$ & $t=1$ if $\bx$ and $\by$ are continuous segments from corpus.\\
   \midrule
   RTD & $\displaystyle \mathcal{L}_{\textrm{\tiny RTD}}=-\sum_{t=1}^{T} \log p(y_t|\hat{\bx})$ & $y_t=\mathbf{1}(\hat{x}_t=x_t)$, $\hat{\bx}$ is corrupted from $\bx$.\\
 \bottomrule
 \end{tabular}
 \begin{tablenotes}
\item[$1$] $\bx=[x_1,x_2,\cdots,x_T]$ denotes a sequence.
\end{tablenotes}
 \end{threeparttable}
\end{table*}

\subsubsection{Language Modeling (LM)}

The most common unsupervised task in NLP is probabilistic language modeling (LM), which is a classic probabilistic density estimation problem.
Although LM is a general concept, in practice, LM often refers in particular to auto-regressive LM or unidirectional LM.

Given a text sequence $\bx_{1:T}=[x_1,x_2,\cdots,x_T ]$, its joint probability $p(x_{1:T})$ can be decomposed as
\begin{align}
p(\bx_{1:T}) = \prod_{t=1}^{T}p(x_t|\bx_{0:t-1}),
\end{align}
where $x_0$ is special token indicating the begin of sequence.

The conditional probability $p(x_t|\bx_{0:t-1})$ can be modeled by a probability distribution over the vocabulary given linguistic context $\bx_{0:t-1}$.
The context $\bx_{0:t-1}$ is modeled by neural encoder $f_{\mathrm{enc}}(\cdot)$, and the conditional probability is
\begin{align}
p(x_t|\bx_{0:t-1}) = g_{\mathrm{LM}}\Big( f_{\mathrm{enc}}(\bx_{0:t-1})\Big),
\end{align}
where $g_{\mathrm{LM}}(\cdot)$ is prediction layer.

Given a huge corpus, we can train the entire network with maximum likelihood estimation (MLE).

A drawback of unidirectional LM is that the representation of each token encodes only the leftward context tokens and itself. However, better contextual
representations of text should encode contextual information from both directions. An improved solution is bidirectional LM (BiLM), which consists of two unidirectional LMs:  a forward left-to-right LM and a backward right-to-left LM.
For BiLM, \citet{DBLP:conf/emnlp/BaevskiELZA19} proposed a two-tower model that the forward tower operates the left-to-right LM and the backward tower operates the right-to-left LM.

\subsubsection{Masked Language Modeling (MLM)}

Masked language modeling (MLM) is first proposed by \citet{doi:10.1177/107769905303000401} in the literature, who referred to this as a Cloze task. \citet{devlin2019bert} adapted this task as a novel pre-training task to overcome the drawback of the standard unidirectional LM. Loosely speaking, MLM first masks out some tokens from the input sentences and then trains the model to predict the masked tokens by the rest of the tokens. However, this pre-training method will create a mismatch between the pre-training phase and the fine-tuning phase because the mask token does not appear during the fine-tuning phase. Empirically, to deal with this issue, \citet{devlin2019bert} used a special \texttt{[MASK]} token 80\% of the time, a random token 10\% of the time and the original token 10\% of the time to perform masking.

\paragraph{Sequence-to-Sequence MLM (Seq2Seq MLM)}
MLM is usually solved as classification problem.
We feed the masked sequences to a neural encoder whose output vectors are further fed into a softmax classifier to predict the masked token.
Alternatively, we can use encoder-decoder (aka. sequence-to-sequence) architecture for MLM, in which the encoder is fed a masked sequence, and the decoder sequentially produces the masked tokens in auto-regression fashion.
We refer to this kind of MLM as sequence-to-sequence MLM (Seq2Seq MLM), which is used in MASS~\cite{DBLP:conf/icml/SongTQLL19} and T5~\cite{raffel2019t5}. Seq2Seq MLM can benefit the Seq2Seq-style downstream tasks, such as question answering, summarization, and machine translation.


\paragraph{Enhanced Masked Language Modeling (E-MLM)}
Concurrently, there are multiple research proposing different enhanced versions of MLM to further improve on BERT.
Instead of static masking, RoBERTa~\cite{liu2019roberta} improves BERT by dynamic masking.

UniLM~\cite{DBLP:conf/nips/00040WWLWGZH19,bao2020unilmv2} extends the task of mask prediction on three types of language modeling tasks: unidirectional, bidirectional, and sequence-to-sequence prediction.
XLM~\cite{DBLP:conf/nips/ConneauL19} performs MLM on a concatenation of parallel bilingual sentence pairs, called \textit{Translation Language Modeling} (TLM).
SpanBERT~\cite{joshi2019spanbert} replaces MLM with \textit{Random Contiguous Words Masking} and \textit{Span Boundary Objective} (SBO) to integrate structure information into pre-training, which requires the system to predict masked spans based on span boundaries. Besides, StructBERT~\cite{wang2020structbert} introduces the \textit{Span Order Recovery} task to further incorporate language structures.

Another way to enrich MLM is to incorporate external knowledge (see Section \ref{sec:ptms-knowledge}).

\subsubsection{Permuted Language Modeling (PLM)}
Despite the wide use of the MLM task in pre-training, \citet{yang2019xlnet} claimed that some special tokens used in the pre-training of MLM, like \texttt{[MASK]}, are absent when the model is applied on downstream tasks, leading to a gap between pre-training and fine-tuning. To overcome this issue, Permuted Language Modeling (PLM)~\cite{yang2019xlnet} is a pre-training objective to replace MLM. In short, PLM is a language modeling task on a random permutation of input sequences. A permutation is randomly sampled from all possible permutations. Then some of the tokens in the permuted sequence are chosen as the target, and the model is trained to predict these targets, depending on the rest of the tokens and the natural positions of targets. Note that this permutation does not affect the natural positions of sequences and only defines the order of token predictions. In practice, only the last few tokens in the permuted sequences are predicted, due to the slow convergence. And a special two-stream self-attention is introduced for target-aware representations.

\subsubsection{Denoising Autoencoder (DAE)}

Denoising autoencoder (DAE) takes a partially corrupted input and aims to recover the original undistorted input. Specific to language, a sequence-to-sequence model, such as the standard Transformer, is used to reconstruct the original text. There are several ways to corrupt text~\cite{lewis2019bart}:

(1) \textit{Token Masking:} Randomly sampling tokens from the input and replacing them with \texttt{[MASK]} elements.

(2) \textit{Token Deletion:} Randomly deleting tokens from the input. Different from token masking, the model needs to decide the positions of missing inputs.

(3) \textit{Text Infilling:} Like SpanBERT, a number of text spans are sampled and replaced with a single \texttt{[MASK]} token. Each span length is drawn from a Poisson distribution ($\lambda = 3$). The model needs to predict how many tokens are missing from a span.

(4) \textit{Sentence Permutation:} Dividing a document into sentences based on full stops and shuffling these sentences in random order.

(5) \textit{Document Rotation:} Selecting a token uniformly at random and rotating the document so that it begins with that token. The model needs to identify the real start position of the document.

\subsubsection{Contrastive Learning (CTL)}

Contrastive learning~\cite{saunshi2019theoretical}
assumes some observed pairs of text that are more semantically similar than randomly sampled
text. A score function $s(x,y)$ for text pair $(x,y)$ is learned to minimize the objective function:
\begin{align}
\mathcal{L}_{\textrm{\tiny CTL}}=\mathbb{E}_{x,y^{+},y^{-}}\Big[-\log \frac{ \exp\big(s(x,y^{+})\big)}
{\exp\big(s(x,y^{+})\big)+\exp\big(s(x,y^{-})\big)
}
\Big],
\end{align}
where $(x, y^{+})$ are a similar pair and $y^{-}$
is presumably dissimilar to $x$. $y^{+}$ and $y^{-}$ are typically called positive and negative sample. The score function $s(x,y)$ is often computed by a learnable neural encoder in two ways: $s(x,y)=f_{\mathrm{enc}(x)}\tran f_{\mathrm{enc}(y)}$ or $s(x,y)=f_{\mathrm{enc}}(x\oplus y)$.

The idea behind CTL is ``learning by comparison''.
Compared to LM, CTL usually has less computational complexity and therefore is desirable alternative training criteria for PTMs.

\citet{DBLP:journals/jmlr/CollobertWBKKK11} proposed \textit{pairwise ranking} task to distinguish real and fake phrases. The model needs to predict a higher score for a legal phrase than an incorrect phrase obtained by replacing its central word with a random word.
\citet{mnih2013learning} trained word embeddings efficiently with Noise-Contrastive Estimation (NCE)~\cite{gutmann2010noise}, which trains a binary classifier to distinguish real and fake samples. The idea of NCE is also used in the well-known word2vec embedding~\cite{mikolov2013word2vec}.

%


We briefly describe some recently proposed CTL tasks in the following paragraphs.

\paragraph{Deep InfoMax (DIM)}
Deep InfoMax (DIM)~\cite{DBLP:conf/iclr/HjelmFLGBTB19} is originally proposed for images, which improves the quality of the representation by maximizing the mutual information between an image
representation and local regions of the image.

\citet{kong2019mutual} applied DIM to language representation learning.
The global representation of a sequence $x$ is defined to be the hidden state of the first token (assumed to be a special start of sentence symbol) output by contextual encoder $f_{\mathrm{enc}}(\bx)$.
The objective of DIM is to assign a higher score for
$f_{\mathrm{enc}}(\bx_{i:j})\tran f_{\mathrm{enc}}(\hat{\bx}_{i:j})$ than $f_{\mathrm{enc}}(\tilde{\bx}_{i:j})\tran f_{\mathrm{enc}}(\hat{\bx}_{i:j})$,
where $\bx_{i:j}$ denotes an n-gram\footnote{$n$ is drawn from a Gaussian distribution
$\mathcal{N}(5, 1)$ clipped at 1 (minimum length) and 10 (maximum length).} span from $i$ to $j$ in $\bx$, $\hat{\bx}_{i:j}$ denotes a sentence masked at position $i$ to $j$,  and $\tilde{\bx}_{i:j}$ denotes a randomly-sampled  negative n-gram from corpus.

\paragraph{Replaced Token Detection (RTD)}

Replaced Token Detection (RTD) is the same as NCE but predicts whether a token is replaced given its surrounding context.

CBOW with negative sampling (CBOW-NS)~\cite{mikolov2013word2vec} can be viewed as a simple version of RTD, in which the negative samples are randomly sampled from vocabulary with simple proposal distribution.

ELECTRA~\cite{clark2020electra} improves RTD by utilizing a generator to replacing some tokens of a sequence. A generator $G$ and a discriminator $D$ are trained following a two-stage procedure: (1) Train only the generator with MLM task for $n_1$ steps; (2) Initialize the weights of the discriminator with the weights of the generator. Then train the discriminator with a discriminative task for $n_2$ steps, keeping $G$ frozen. Here the discriminative task indicates justifying whether the input token has been replaced by $G$ or not.
The generator is thrown after pre-training, and only the discriminator will be fine-tuned on downstream tasks.

RTD is also an alternative solution for the mismatch problem. The network sees \texttt{[MASK]} during pre-training but not when being fine-tuned in downstream tasks.

Similarly, WKLM~\cite{xiong2019pretrain} replaces words on the entity-level instead of token-level. Concretely, WKLM replaces entity mentions with names of other entities of the same type and train the models to distinguish whether the entity has been replaced.

\tikzstyle{leaf}=[draw=hiddendraw,
    rounded corners,minimum height=1.2em,
    fill=hidden-orange!40,text opacity=1,    align=center,
    fill opacity=.5,  text=black,align=left,font=\scriptsize,
inner xsep=3pt,
inner ysep=1pt,
]

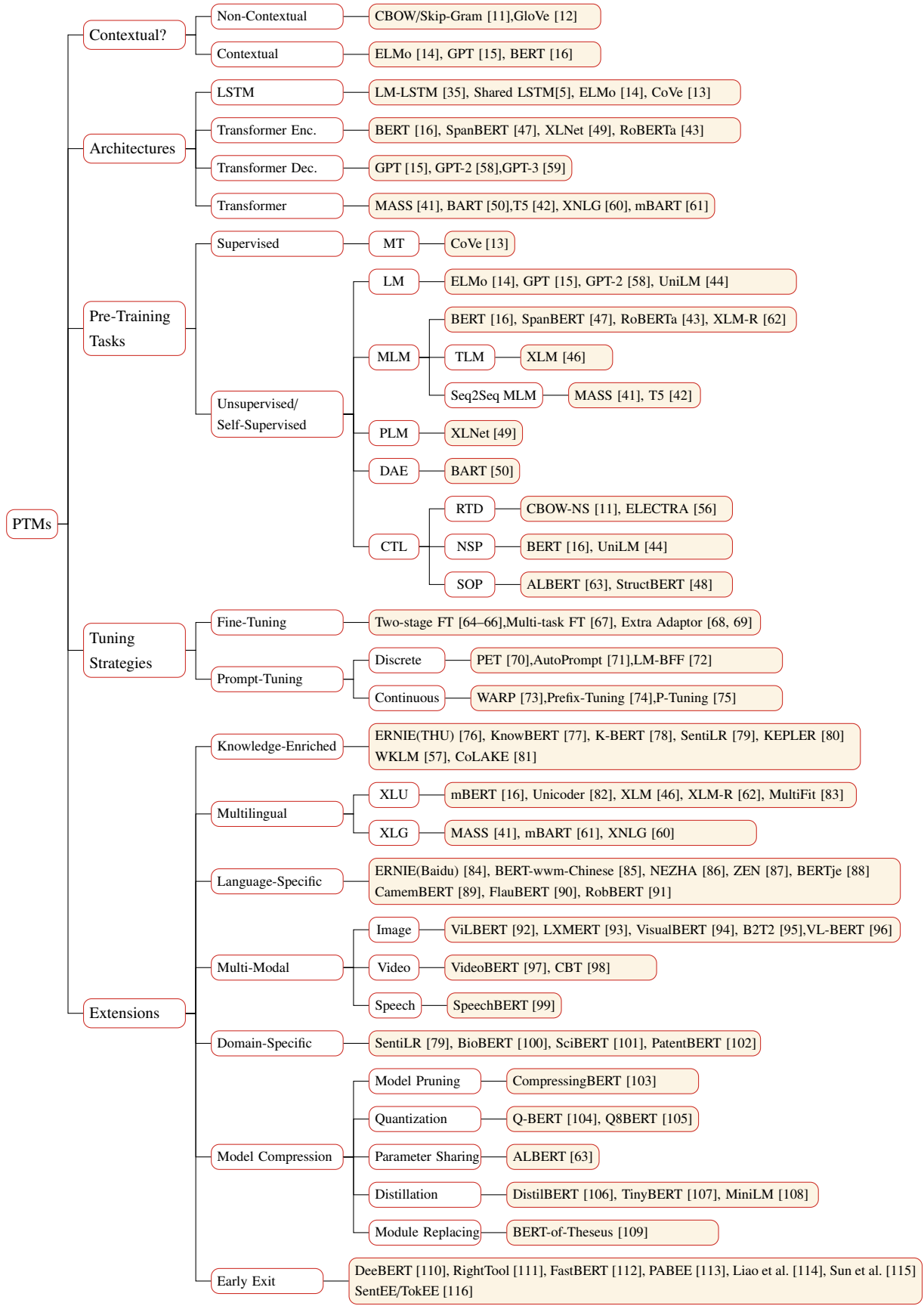
\begin{figure*}[thp]
  \centering
\begin{forest}
  forked edges,
  for tree={
  grow=east,
  reversed=true,
  anchor=base west,
  parent anchor=east,
  child anchor=west,
  base=left,
  font=\footnotesize,
  rectangle,
  draw=hiddendraw,
  rounded corners,align=left,
  minimum width=2.5em,
  minimum height=1.2em,
s sep=6pt,
inner xsep=3pt,
inner ysep=1pt,
  },
  where level=1{text width=4.5em}{},
  where level=2{text width=6em,font=\scriptsize}{},
  where level=3{font=\scriptsize}{},
  where level=4{font=\scriptsize}{},
  where level=5{font=\scriptsize}{},
  [PTMs
    [Contextual?
        [Non-Contextual
           [CBOW/Skip-Gram~\cite{mikolov2013word2vec}{,}GloVe~\cite{DBLP:conf/emnlp/PenningtonSM14}
           ,leaf,text width=11em]
        ]
        [Contextual
            [ELMo~\cite{peters2018elmo}{,}
            GPT~\cite{radford2018improving}{,} BERT~\cite{devlin2019bert}
            ,leaf,text width=11em]
        ]
    ]
    [Architectures
     [LSTM
        [LM-LSTM~\cite{dai2015semi}{,} Shared LSTM\cite{liu2016recurrent}{,}
        ELMo~\cite{peters2018elmo}{,}
        CoVe~\cite{mccan2017learn},leaf,text width=18em]
     ]
    [Transformer Enc.
        [BERT~\cite{devlin2019bert}{,}
        SpanBERT~\cite{joshi2019spanbert}{,}
        XLNet~\cite{yang2019xlnet}{,}
        RoBERTa~\cite{liu2019roberta}
        ,leaf,text width=18em]
    ]
    [Transformer Dec.
        [GPT~\cite{radford2018improving}{,}
        GPT-2~\cite{radford2019language}{,}GPT-3~\cite{Brown2020GPT3}
        ,leaf]
    ]
    [Transformer
        [MASS~\cite{DBLP:conf/icml/SongTQLL19}{,}
        BART~\cite{lewis2019bart}{,}T5~\cite{raffel2019t5}{,}
        XNLG~\cite{chi2019cross}{,}
        mBART~\cite{liu2020multilingual}
        ,leaf]
    ]
    ]
    [Pre-Training\\Tasks
      [Supervised
        [MT
            [CoVe~\cite{mccan2017learn}
            ,leaf]
        ]
      ]
      [Unsupervised/\\Self-Supervised
        [LM
            [ELMo~\cite{peters2018elmo}{,}
            GPT~\cite{radford2018improving}{,}
            GPT-2~\cite{radford2019language}{,}
            UniLM~\cite{DBLP:conf/nips/00040WWLWGZH19}
            ,leaf,text width=17em]
        ]
        [MLM
            [BERT~\cite{devlin2019bert}{,}
            SpanBERT~\cite{joshi2019spanbert}{,}
            RoBERTa~\cite{liu2019roberta}{,}
            XLM-R~\cite{conneau2019unsupervised}
            ,leaf,text width=17em]
            [TLM
                [XLM~\cite{DBLP:conf/nips/ConneauL19}
                ,leaf,text width=4em
                ]
            ]
            [Seq2Seq MLM
              [MASS~\cite{DBLP:conf/icml/SongTQLL19}{,}
                T5~\cite{raffel2019t5}
                ,leaf,text width=6em
              ]
            ]
        ]
        [PLM
            [XLNet~\cite{yang2019xlnet},leaf]
        ]
        [DAE
            [BART~\cite{lewis2019bart},leaf]
        ]
        [CTL
            [RTD
                [CBOW-NS~\cite{mikolov2013word2vec}{,}
                ELECTRA~\cite{clark2020electra},leaf,text width=10em]
            ]
            [NSP
                [BERT~\cite{devlin2019bert}{,}
                UniLM~\cite{DBLP:conf/nips/00040WWLWGZH19}
                ,leaf,text width=10em]
            ]
            [SOP
                [ALBERT~\cite{lan2019albert}{,}
                StructBERT~\cite{wang2020structbert}
                ,leaf,text width=10em]
            ]
         ]
      ]
    ]
    [Tuning\\Strategies
      [Fine-Tuning
        [Two-stage FT~\cite{sun2019fine,li2019story,gururangan2020don}{,}Multi-task FT~\cite{liu2019multi}{,}
        Extra Adaptor~\cite{stickland2019bert,houlsby2019parameter}
        ,leaf,text width=19em]
      ]
      [Prompt-Tuning
        [Discrete,text width=3.2em
            [PET~\cite{Schick2021PET}{,}AutoPrompt~\cite{Shin2020AutoPrompt}{,}LM-BFF~\cite{Gao2020LMBFF}
            ,leaf,text width=15em]
        ]
        [Continuous,text width=3.2em
            [WARP~\cite{Hambardzumyan2021WARP}{,}Prefix-Tuning~\cite{Li2021Prefix}{,}P-Tuning~\cite{Liu2021PTuning}
            ,leaf,text width=15em]
        ]
      ]
    ]
    [Extensions
      [Knowledge-Enriched
        [ERNIE(THU)~\cite{zhang2019ernie}{,}
        KnowBERT~\cite{peters2019knowledge}{,}
        K-BERT~\cite{liu2019kbert}{,}
        SentiLR~\cite{ke2019sentilr}{,}
        KEPLER~\cite{wang2019kepler}\\
        WKLM~\cite{xiong2019pretrain}{,}
        CoLAKE~\cite{Sun2020CoLAKE},leaf,text width=24em]
      ]
      [Multilingual
        [XLU
            [mBERT~\cite{devlin2019bert}{,}
              Unicoder~\cite{huang2019unicoder}{,}
              XLM~\cite{DBLP:conf/nips/ConneauL19}{,}
              XLM-R~\cite{conneau2019unsupervised}{,}
              MultiFit~\cite{DBLP:conf/emnlp/EisenschlosRCKG19}
              ,leaf,text width=20em]
         ]
         [XLG
              [
              MASS~\cite{DBLP:conf/icml/SongTQLL19}{,}
              mBART~\cite{liu2020multilingual}{,}
              XNLG~\cite{chi2019cross}
              ,leaf,text width=15em]L
         ]
      ]
      [Language-Specific
        [ERNIE(Baidu)~\cite{sun2019ernie}{,}
        BERT-wwm-Chinese~\cite{cui2019pretraining}{,} NEZHA~\cite{wei2019nezha}{,}
        ZEN~\cite{diao2019zen}{,}
        BERTje~\cite{vries2019bertje}\\
        CamemBERT~\cite{martin2019camenbert}{,}
        FlauBERT~\cite{le2019flaubert}{,}
        RobBERT~\cite{delobelle2020robbert},leaf,text width=26em]
      ]
      [Multi-Modal
        [Image
            [ViLBERT~\cite{lu2019vilbert}{,}
                LXMERT~\cite{DBLP:conf/emnlp/TanB19}{,}
                VisualBERT~\cite{DBLP:journals/corr/abs-1908-03557}{,}
                B2T2~\cite{DBLP:conf/emnlp/AlbertiLCR19}{,}VL-BERT~\cite{su2020vl-bert}
            ,leaf,text width=22.2em]
         ]
         [Video
            [VideoBERT~\cite{sun2019videobert}{,}
                CBT~\cite{DBLP:journals/corr/abs-1906-05743}
            ,leaf,text width=10em]
         ]
         [Speech
            [SpeechBERT~\cite{DBLP:journals/corr/abs-1910-11559}
            ,leaf]
         ]
      ]
      [Domain-Specific
        [SentiLR~\cite{ke2019sentilr}{,}
        BioBERT~\cite{lee2019biobert}{,}
        SciBERT~\cite{beltagy2019scibert}{,}
        PatentBERT~\cite{lee2019patentbert},leaf,text width=19em]
      ]
      [Model Compression
        [Model Pruning,text width=5em
          [CompressingBERT~\cite{gordon2020compressing},leaf,text width=9em]
        ]
        [Quantization,text width=5em
          [Q-BERT~\cite{shen2020q}{,} Q8BERT~\cite{zafrir2019q8bert},leaf, text width=9em]
        ]
        [Parameter Sharing,text width=5em
          [ALBERT~\cite{lan2019albert},leaf]
        ]
        [Distillation,text width=5em
          [DistilBERT~\cite{sanh2019distilbert}{,}
          TinyBERT~\cite{jiao2019tinybert}{,} MiniLM~\cite{wang2020minilm}
          ,leaf,text width=15em
          ]
        ]
        [Module Replacing,text width=5em
          [BERT-of-Theseus~\cite{xu2020bert}
          ,leaf,text width=10em]
        ]
      ]
      [Early Exit,text width=5em
          [DeeBERT~\cite{xin-etal-2020-deebert}{,}
          RightTool~\cite{DBLP:conf/acl/SchwartzSSDS20}{,}
          FastBERT~\cite{liu2020fastbert}{,}
          PABEE~\cite{zhou2020bert}{,}
          \citet{Liao2021Global}{,}
          \citet{Sun2021Early}\\
          SentEE/TokEE~\cite{li2021accelerating}
          ,leaf]
      ]
    ]
  ]
\end{forest}
\caption{Taxonomy of PTMs with Representative Examples}
\label{taxonomy_of_PTMs}
\end{figure*}

\begin{table*}[t]\small
\scriptsize\centering
\begin{threeparttable}\caption{List of Representative PTMs}\label{tab:listofPTMs}
\doublerulesep 0.1pt
\tabcolsep 4.5pt 
\begin{tabular}{l|p{0.1\linewidth}p{0.05\linewidth}@{\hskip 2.5em}p{0.12\linewidth}@{\hskip 2em}P{0.22\linewidth}@{\hskip 2em}lcc}
\toprule
  PTMs & Architecture\tnote{$\dagger$} &  Input & Pre-Training Task & Corpus & Params & GLUE\tnote{$\ddagger$} & FT?\tnote{$\sharp$}\\
\midrule
 ELMo~\cite{peters2018elmo} & LSTM & Text & BiLM & WikiText-103&&& No\\
 GPT~\cite{radford2018improving} & Transformer Dec. & Text & LM & BookCorpus & 117M & 72.8 & Yes\\
 GPT-2~\cite{radford2019language} & Transformer Dec. & Text & LM & WebText & 117M $\sim$ 1542M &  & No\\
 BERT~\cite{devlin2019bert} & Transformer Enc. & Text & MLM \& NSP & WikiEn+BookCorpus & 110M $\sim$ 340M & 81.9\tnote{$*$} & Yes\\
 InfoWord~\cite{kong2019mutual} & Transformer Enc. & Text & DIM+MLM & WikiEn+BookCorpus  & =BERT & 81.1\tnote{$*$} & Yes\\
 RoBERTa~\cite{liu2019roberta} & Transformer Enc. & Text & MLM & BookCorpus+CC-News+OpenWebText+ STORIES & 355M & 88.5 & Yes\\
 XLNet~\cite{yang2019xlnet} & Two-Stream Transformer Enc. & Text & PLM & WikiEn+ BookCorpus+Giga5 +ClueWeb+Common Crawl & $\approx$BERT & 90.5\tnote{$\mathsection$} & Yes\\
 ELECTRA~\cite{clark2020electra} & Transformer Enc. & Text & RTD+MLM & same to XLNet  &  335M & 88.6 & Yes\\
 UniLM~\cite{DBLP:conf/nips/00040WWLWGZH19}& Transformer Enc. & Text & MLM\tnote{$\diamond$}+ NSP & WikiEn+BookCorpus & 340M & 80.8 & Yes\\
 MASS~\cite{DBLP:conf/icml/SongTQLL19} & Transformer & Text & Seq2Seq MLM  & *Task-dependent &  &  & Yes\\
 BART~\cite{lewis2019bart}& Transformer & Text & DAE & same to RoBERTa & 110\% of BERT & 88.4\tnote{$*$} & Yes\\
 T5~\cite{raffel2019t5} & Transformer & Text & Seq2Seq MLM & Colossal Clean Crawled Corpus (C4) & 220M $\sim$ 11B & 89.7\tnote{$*$} & Yes\\
 \midrule
 ERNIE(THU)~\cite{zhang2019ernie} & Transformer Enc. & Text+Entities & MLM+NSP+dEA & WikiEn + Wikidata & 114M & 79.6 & Yes \\
 KnowBERT~\cite{peters2019knowledge} & Transformer Enc. & Text & MLM+NSP+EL & WikiEn + WordNet/Wiki & 253M $\sim$ 523M& & Yes \\
 K-BERT~\cite{liu2019kbert} & Transformer Enc. & Text+Triples & MLM+NSP & WikiZh + WebtextZh + CN-DBpedia + HowNet + MedicalKG & $=$BERT & & Yes \\
 KEPLER~\cite{wang2019kepler} & Transformer Enc. & Text & MLM+KE & WikiEn + Wikidata/WordNet &  & & Yes \\
 WKLM~\cite{xiong2019pretrain} & Transformer Enc. & Text & MLM+ERD & WikiEn + Wikidata & $=$BERT & & Yes\\
 CoLAKE~\cite{Sun2020CoLAKE} & Transformer Enc. & Text+Triples & MLM & WikiEn + Wikidata & =RoBERTa & 86.3 & Yes\\
 \bottomrule
 \end{tabular}
\begin{tablenotes}
\item[$\dagger$] ``Transformer Enc.'' and ``Transformer Dec.'' mean the encoder and decoder part of the standard Transformer architecture respectively. Their difference is that the decoder part uses masked self-attention with triangular matrix to prevent tokens from attending their future (right) positions. ``Transformer'' means the standard encoder-decoder architecture.
\item[$\ddagger$] the averaged score on 9 tasks of GLUE benchmark (see Section \ref{sec:app-glue}).
\item[$*$] without WNLI task.
\item[$\mathsection$] indicates ensemble result.
\item[$\sharp$] means whether is model usually used in fine-tuning fashion.
 \item[$\diamond$] The MLM of UniLM is built on three versions of LMs: Unidirectional LM, Bidirectional LM,  and Sequence-to-Sequence LM.
\end{tablenotes}
 \end{threeparttable}
 \end{table*}

\paragraph{Next Sentence Prediction (NSP)}

Punctuations are the natural separators of text data. So, it is reasonable to construct pre-training methods by utilizing them. Next Sentence Prediction (NSP)~\cite{devlin2019bert} is just a great example of this. As its name suggests, NSP trains the model to distinguish whether two input sentences are continuous segments from the training corpus. Specifically, when choosing the sentences pair for each pre-training example, 50\% of the time, the second sentence is the actual next sentence of the first one, and 50\% of the time, it is a random sentence from the corpus. By doing so, it is capable to teach the model to understand the relationship between two input sentences and thus benefit downstream tasks that are sensitive to this information, such as Question Answering and Natural Language Inference.


However, the necessity of the NSP task has been questioned by subsequent work \cite{joshi2019spanbert,yang2019xlnet,liu2019roberta,lan2019albert}. \citet{yang2019xlnet} found the impact of the NSP task unreliable, while \citet{joshi2019spanbert} found that single-sentence training without the NSP loss is superior to sentence-pair training with the NSP loss. Moreover, \citet{liu2019roberta} conducted a further analysis for the NSP task, which shows that when training with blocks of text from a single document, removing the NSP loss matches or slightly improves performance on downstream tasks.

\paragraph{Sentence Order Prediction (SOP)}

To better model inter-sentence coherence, ALBERT \cite{lan2019albert} replaces the NSP loss with a sentence order prediction (SOP) loss.
As conjectured in \citet{lan2019albert}, NSP conflates topic prediction and coherence prediction in a single task. Thus, the model is allowed to make predictions merely rely on the easier task, topic prediction.
Different from NSP, SOP uses two consecutive segments from the same document as positive examples, and the same two consecutive segments but with their order swapped as negative examples.
As a result, ALBERT consistently outperforms BERT on various downstream tasks.

StructBERT~\cite{wang2020structbert} and BERTje~\cite{vries2019bertje} also take SOP as their self-supervised learning task.

\subsubsection{Others}

Apart from the above tasks, there are many other auxiliary pre-training tasks designated  to incorporate factual knowledge  (see Section \ref{sec:ptms-knowledge}), improve cross-lingual tasks (see Section \ref{sec:ptms-multilingual}), multi-modal applications (see Section \ref{sec:ptms-mutimodel}), or other specific tasks (see Section \ref{sec:ptms-domain}).


\subsection{Taxonomy of PTMs}
\label{sec:taxonomy}

To clarify the relations of existing PTMs for NLP, we build the taxonomy of PTMs, which categorizes existing PTMs from four different perspectives:
\begin{enumerate}
  \item \textit{Representation Type}: According to the representation used for downstream tasks, we can divide PTMs into non-contextual and contextual models.
  \item \textit{Architectures}: The backbone network used by PTMs, including LSTM, Transformer encoder, Transformer decoder, and the full Transformer architecture. ``Transformer'' means the standard encoder-decoder architecture. ``Transformer encoder'' and ``Transformer decoder'' mean the encoder and decoder part of the standard Transformer architecture, respectively. Their difference is that the decoder part uses masked self-attention with a triangular matrix to prevent tokens from attending their future (right) positions.
  \item \textit{Pre-Training Task Types}: The type of pre-training tasks used by PTMs. We have discussed them in Section \ref{sec:pre-training-tasks}.
  \item \textit{Extensions}: PTMs designed for various scenarios, including knowledge-enriched PTMs, multilingual or language-specific PTMs, multi-model PTMs, domain-specific PTMs and compressed PTMs.
      We will particularly introduce these extensions in Section \ref{sec:extension}.
\end{enumerate}

Figure~\ref{taxonomy_of_PTMs} shows the taxonomy as well as some corresponding representative PTMs.
Besides, Table~\ref{tab:listofPTMs} distinguishes some representative PTMs in more detail.

\todo[inline]{add section for Training Skill}

\subsection{Model Analysis}
\label{sec:model-analysis}

Due to the great success of PTMs, it is important to understand what kinds of knowledge are captured by them, and how to induce knowledge from them. There is a wide range of literature analyzing linguistic knowledge and world knowledge stored in pre-trained non-contextual and contextual embeddings.

\subsubsection{Non-Contextual Embeddings}
Static word embeddings are first probed for kinds of knowledge. \citet{mikolov2013ling} found that word representations learned by neural network language models are able to capture linguistic regularities in language, and the relationship between words can be characterized by a relation-specific vector offset. Further analogy experiments \cite{mikolov2013word2vec} demonstrated that word vectors produced by skip-gram model can capture both syntactic and semantic word relationships, such as vec(``China'') $-$ vec(``Beijing'') $\approx$ vec(``Japan'') $-$ vec(``Tokyo''). Besides, they find compositionality property of word vectors, for example, vec(``Germany'') $+$ vec(``capital'') is close to vec(``Berlin''). Inspired by these work, \citet{rubinstein2015how} found that distributional word representations are good at predicting taxonomic properties (e.g., dog is an animal) but fail to learn attributive properties (e.g., swan is white). Similarly, \citet{gupta2015distributional} showed that word2vec embeddings implicitly encode referential attributes of entities. The distributed word vectors, along with a simple supervised model, can learn to predict numeric and binary attributes of entities with a reasonable degree of accuracy.

\subsubsection{Contextual Embeddings}
A large number of studies have probed and induced different types of knowledge in contextual embeddings. In general, there are two types of knowledge: linguistic knowledge and world knowledge.

\paragraph{Linguistic Knowledge}

A wide range of probing tasks are designed to investigate the linguistic knowledge in PTMs. \citet{tenney2019what,liu2019ling} found that BERT performs well on many syntactic tasks such as part-of-speech tagging and constituent labeling. However, BERT is not good enough at semantic and fine-grained syntactic tasks, compared with simple syntactic tasks.

Besides, \citet{DBLP:conf/acl/TenneyDP19} analyzed the roles of BERT's layers in different tasks and found that BERT solves tasks in a similar order to that in NLP pipelines.
Furthermore, knowledge of subject-verb agreement~\cite{goldberg2019assessing} and semantic roles~\cite{ettinger2020what} are also confirmed to exist in BERT. Besides, \citet{hewitt2019structural,jawahar2019what,kim2020are} proposed several methods to extract dependency trees and constituency trees from BERT, which proved the BERT's ability to encode syntax structure.
\citet{reif2019visualizing} explored the geometry of internal representations in BERT and find some evidence: 1)  linguistic features seem to be represented in separate semantic and
syntactic subspaces; 2) attention matrices contain grammatical representations; 3) BERT distinguishes word senses at a very fine level.

\paragraph{World Knowledge}

Besides linguistic knowledge, PTMs may also store world knowledge presented in the training data. A straightforward method of probing world knowledge is to query BERT with ``fill-in-the-blank'' cloze statements, for example, ``Dante was born in \texttt{[MASK]}''.  \citet{petroni2019language} constructed LAMA (Language Model Analysis) task by manually creating single-token cloze statements (queries) from several knowledge sources. Their experiments show that BERT contains world knowledge competitive with traditional information extraction methods. Since the simplicity of query generation procedure in LAMA, \citet{jiang2019how} argued that LAMA just measures a lower bound for what language models know and propose more advanced methods to generate more efficient queries. Despite the surprising findings of LAMA, it has also been questioned by subsequent work~\cite{porner2019bert,kassner2019negated}. Similarly, several studies induce relational knowledge \cite{bouraoui2019inducing} and commonsense knowledge \cite{davison2019commonsense} from BERT for downstream tasks.

\section{Extensions of PTMs}
\label{sec:extension}

\subsection{Knowledge-Enriched PTMs}
\label{sec:ptms-knowledge}



PTMs usually learn universal language representation from general-purpose large-scale text corpora but lack domain-specific knowledge. Incorporating domain knowledge from external knowledge bases into PTM has been shown to be effective.
The external knowledge ranges from linguistic~\cite{Lauscher2019informing,ke2019sentilr,peters2019knowledge,wang2020kadapter}, semantic~\cite{levine2019sensebert}, commonsense~\cite{guan2020knowledge}, factual~\cite{zhang2019ernie,peters2019knowledge,liu2019kbert,xiong2019pretrain,wang2019kepler}, to domain-specific knowledge~\cite{he2019integrating,liu2019kbert}.

On the one hand, external knowledge can be injected during pre-training.
Early studies~\cite{wang2014knowledge,zhong2015aligning,xie2016representation,xu2017knowledge} focused on learning knowledge graph embeddings and word embedding jointly.
Since BERT, some auxiliary pre-training tasks are designed to incorporate external knowledge into deep PTMs.
LIBERT~\cite{Lauscher2019informing} (linguistically-informed BERT) incorporates linguistic knowledge via an additional linguistic constraint task. \citet{ke2019sentilr} integrated sentiment polarity of each word to extend the MLM to Label-Aware MLM (LA-MLM). As a result, their proposed model, SentiLR, achieves state-of-the-art performance on several sentence- and aspect-level sentiment classification tasks. \citet{levine2019sensebert} proposed SenseBERT, which is pre-trained to predict not only the masked tokens but also their supersenses in WordNet. ERNIE(THU)~\cite{zhang2019ernie} integrates entity embeddings pre-trained on a knowledge graph with corresponding entity mentions in the text to enhance the text representation. Similarly, KnowBERT~\cite{peters2019knowledge} trains BERT jointly with an entity linking model to incorporate entity representation in an end-to-end fashion. \citet{wang2019kepler} proposed KEPLER, which jointly optimizes knowledge embedding and language modeling objectives. These work inject structure information of knowledge graph via entity embedding. In contrast, K-BERT~\cite{liu2019kbert} explicitly injects related triples extracted from KG into the sentence to obtain an extended tree-form input for BERT. CoLAKE~\cite{Sun2020CoLAKE} integrates knowledge context and language context into a unified graph, which is then pre-trained with MLM to obtain contextualized representation for both knowledge and language. Moreover, \citet{xiong2019pretrain} adopted entity replacement identification to encourage the model to be more aware of factual knowledge. However, most of these methods update the parameters of PTMs when injecting knowledge, which may suffer from catastrophic forgetting when injecting multiple kinds of knowledge. To address this, K-Adapter~\cite{wang2020kadapter} injects multiple kinds of knowledge by training different adapters independently for different pre-training tasks, which allows continual knowledge infusion.

On the other hand, one can incorporate external knowledge into pre-trained models without retraining them from scratch. As an example, K-BERT~\cite{liu2019kbert} allows injecting factual knowledge during fine-tuning on downstream tasks. \citet{guan2020knowledge} employed commonsense knowledge bases, ConceptNet and ATOMIC, to enhance GPT-2 for story generation. \citet{yang2019enhancing} proposed a knowledge-text fusion model to acquire related linguistic and factual knowledge for machine reading comprehension.

Besides, \citet{logan2019barack} and \citet{hayashi2019latent} extended language model to \textit{knowledge graph language model} (KGLM) and \textit{latent relation language model} (LRLM) respectively, both of which allow prediction conditioned on knowledge graph. These novel KG-conditioned language models show potential for pre-training.

\subsection{Multilingual and Language-Specific PTMs}
\label{sec:ptms-multilingual}

\subsubsection{Multilingual PTMs}

Learning multilingual text representations shared across languages plays an important role in many cross-lingual NLP tasks.

\paragraph{Cross-Lingual Language Understanding (XLU)}

Most of the early works focus on learning multilingual word embedding~\cite{faruqui2014improving,luong2015bilingual,singla2018multi}, which represents text from multiple languages in a single semantic space. However, these methods usually need (weak) alignment between languages.


Multilingual BERT\footnote{https://github.com/google-research/bert/blob/master/multilingual.md}
(mBERT) is pre-trained by MLM with the shared vocabulary and weights on Wikipedia
text from the top 104 languages. Each training sample is a monolingual document, and there are no cross-lingual objectives specifically designed nor any cross-lingual data. Even so, mBERT
performs cross-lingual generalization surprisingly
well~\cite{DBLP:journals/corr/abs-1906-01502}. \citet{K2020Cross-Lingual} showed that the lexical overlap between languages plays a negligible role in cross-lingual success.

XLM~\cite{DBLP:conf/nips/ConneauL19} improves mBERT by incorporating a cross-lingual task, translation language modeling (TLM), which performs MLM on a concatenation of parallel bilingual sentence pairs.
Unicoder~\cite{huang2019unicoder} further
propose three new cross-lingual pre-training tasks, including cross-lingual word recovery, cross-lingual paraphrase
classification and cross-lingual masked language
model (XMLM).

XLM-RoBERTa (XLM-R)~\cite{conneau2019unsupervised} is a scaled multilingual encoder pre-trained on a significantly increased amount of training data, 2.5TB clean CommonCrawl data in 100 different languages. The pre-training task of XLM-RoBERTa is monolingual MLM only.
XLM-R achieves state-of-the-arts results on multiple cross-lingual benchmarks, including XNLI, MLQA, and NER.

\paragraph{Cross-Lingual Language Generation (XLG)}

Multilingual generation is a kind of tasks to generate text
with different languages from the input language, such as machine translation and
cross-lingual abstractive summarization.

Different from the PTMs for multilingual classification, the PTMs for multilingual generation usually needs to pre-train both the encoder and decoder jointly, rather than only focusing on the encoder.

MASS~\cite{DBLP:conf/icml/SongTQLL19} pre-trains a Seq2Seq model with monolingual Seq2Seq MLM on multiple languages and achieves
significant improvement for unsupervised NMT.
XNLG~\cite{chi2019cross} performs two-stage pre-training for cross-lingual natural language generation. The first stage pre-trains the encoder with monolingual MLM and Cross-Lingual MLM (XMLM) tasks. The second stage pre-trains the decoder by using monolingual DAE and Cross-Lingual Auto-Encoding (XAE) tasks while keeping the encoder fixed. Experiments show the benefit of XNLG on cross-lingual question generation and cross-lingual abstractive summarization.
mBART~\cite{liu2020multilingual}, a multilingual extension of BART~\cite{lewis2019bart}, pre-trains the encoder and decoder jointly with Seq2Seq denoising auto-encoder (DAE) task on large-scale monolingual corpora across 25 languages. Experiments demonstrate that mBART
produces significant performance gains across a wide variety of machine translation (MT) tasks.

\subsubsection{Language-Specific PTMs}
Although multilingual PTMs perform well on many languages, recent work showed that PTMs trained on a single language significantly outperform the multilingual results~\cite{martin2019camenbert,le2019flaubert,virtanen2019multilingual}.

For Chinese, which does not have explicit word boundaries, modeling larger granularity~\cite{cui2019pretraining,diao2019zen,wei2019nezha} and multi-granularity~\cite{sun2019ernie,sun2019ernie2} word representations have shown great success. \citet{kuratov2019adaptation} used transfer learning techniques to adapt a multilingual PTM to a monolingual PTM for Russian language. In addition, some monolingual PTMs have been released for different languages, such as CamemBERT~\cite{martin2019camenbert} and FlauBERT~\cite{le2019flaubert} for French, FinBERT~\cite{virtanen2019multilingual} for Finnish, BERTje~\cite{vries2019bertje} and  RobBERT~\cite{delobelle2020robbert} for Dutch,
AraBERT~\cite{antoun2020arabert} for Arabic language.

\subsection{Multi-Modal PTMs}
\label{sec:ptms-mutimodel}

Observing the success of PTMs across many NLP tasks, some research has focused on obtaining a cross-modal version of PTMs. A great majority of these models are designed for a general visual and linguistic feature encoding. And these models are pre-trained on some huge corpus of cross-modal data, such as videos with spoken words or images with captions, incorporating extended pre-training tasks to fully utilize the multi-modal feature.
Typically, tasks like \textit{visual-based MLM}, \textit{masked visual-feature modeling} and \textit{visual-linguistic matching} are widely used in multi-modal pre-training, such as VideoBERT~\cite{sun2019videobert}, VisualBERT~\cite{DBLP:journals/corr/abs-1908-03557}, ViLBERT~\cite{lu2019vilbert}.

\subsubsection{Video-Text PTMs}

VideoBERT~\cite{sun2019videobert} and CBT~\cite{DBLP:journals/corr/abs-1906-05743} are joint video and text models. To obtain sequences of visual and linguistic tokens used for pre-training, the videos are pre-processed by CNN-based encoders and off-the-shelf speech recognition techniques, respectively. And a single Transformer encoder is trained on the processed data to learn the vision-language representations for downstream tasks like video caption. Furthermore, UniViLM~\cite{luo2020univilm} proposes to bring in generation tasks to further pre-train the decoder using in downstream tasks.

\subsubsection{Image-Text PTMs}

Besides methods for video-language pre-training, several works introduce PTMs on image-text pairs, aiming to fit downstream tasks like visual question answering(VQA) and visual commonsense reasoning(VCR). Several proposed models adopt two separate encoders for image and text representation independently, such as ViLBERT~\cite{lu2019vilbert} and LXMERT~\cite{DBLP:conf/emnlp/TanB19}. While other methods like VisualBERT~\cite{DBLP:journals/corr/abs-1908-03557}, B2T2~\cite{DBLP:conf/emnlp/AlbertiLCR19}, VL-BERT~\cite{su2020vl-bert}, Unicoder-VL~\cite{li2020unicoder-vl} and UNITER~\cite{DBLP:journals/corr/abs-1909-11740} propose single-stream unified Transformer. Though these model architectures are different, similar pre-training tasks, such as MLM and image-text matching, are introduced in these approaches. And to better exploit visual elements, images are converted into sequences of regions by applying RoI or bounding box retrieval techniques before encoded by pre-trained Transformers.

\subsubsection{Audio-Text PTMs}

Moreover, several methods have explored the chance of PTMs on audio-text pairs, such as SpeechBERT~\cite{DBLP:journals/corr/abs-1910-11559}. This work tries to build an end-to-end Speech Question Answering (SQA) model by encoding audio and text with a single Transformer encoder, which is pre-trained with MLM on speech and text corpus and fine-tuned on Question Answering.

\subsection{Domain-Specific and Task-Specific PTMs}
\label{sec:ptms-domain}

Most publicly available PTMs are trained on general domain corpora such as Wikipedia, which limits their applications to specific domains or tasks. Recently, some studies have proposed PTMs trained on specialty corpora, such as BioBERT~\cite{lee2019biobert} for biomedical text, SciBERT~\cite{beltagy2019scibert} for scientific text, ClinicalBERT~\cite{huang2019clinicalbert,alsentzer2019publicly} for clinical text.

In addition to pre-training a domain-specific PTM, some work attempts to adapt available pre-trained models to target applications, such as biomedical entity normalization~\cite{ji2019bertranking}, patent classification~\cite{lee2019patentbert}, progress notes classification and keyword extraction~\cite{tang2019progress}.

Some task-oriented pre-training tasks were also proposed, such as
sentiment \textit{Label-Aware MLM} in SentiLR~\cite{ke2019sentilr} for sentiment analysis, \textit{Gap Sentence Generation} (GSG)~\cite{zhang2019pegasus} for text summarization, and \textit{Noisy Words Detection} for disfluency detection~\cite{wang2019multi}.



\subsection{Model Compression}
\label{sec:model-compression}

Since PTMs usually consist of at least hundreds of millions of parameters, they are difficult to be deployed on the on-line service in real-life applications and on resource-restricted devices. Model compression ~\cite{bucilua2006model} is a potential approach to reduce the model size and increase computation efficiency.

There are five ways to compress PTMs~\cite{ganesh2020compressing}: (1) \textit{model pruning}, which removes less important parameters, (2) \textit{weight quantization}~\cite{dong2019hawq}, which uses
fewer bits to represent the parameters, (3) \textit{parameter sharing}
across similar model units, (4) \textit{knowledge distillation}~\cite{hinton2015distilling},
which trains a smaller student model that learns from intermediate outputs from the original model
and (5) \textit{module replacing}, which replaces the modules of original PTMs with more compact substitutes.

Table \ref{tab:listofcompressedPTMs} gives a comparison of some representative compressed PTMs.

\begin{table*}[t]\small
\scriptsize\centering
\begin{threeparttable}\caption{Comparison of Compressed PTMs}
\label{tab:listofcompressedPTMs}
\doublerulesep 0.1pt
\tabcolsep 4.5pt 
\begin{tabular}{lllP{0.25\linewidth}@{\hskip 2.5em}l@{\hskip 2em}l@{\hskip 2em}ll}
\toprule
  Method & Type& \#Layer & Loss Function\tnote{$*$} & Speed Up & Params & Source PTM & GLUE\tnote{$\ddagger$}\\
\midrule
BERT$_{\mathrm{BASE}}$~\cite{devlin2019bert} &\multirow{2}*{Baseline}& 12 & $\mathcal{L}_{\textrm{\tiny MLM}}$ + $\mathcal{L}_{\textrm{\tiny NSP}}$ & & 110M & & 79.6 \\
BERT$_{\mathrm{LARGE}}$~\cite{devlin2019bert} & ~ & 24 & $\mathcal{L}_{\textrm{\tiny MLM}}$ + $\mathcal{L}_{\textrm{\tiny NSP}}$ & & 340M &  & 81.9 \\
\midrule
Q-BERT~\cite{shen2020q} & \multirow{2}*{Quantization} & 12 & HAWQ + GWQ & - & & BERT$_{\mathrm{BASE}}$ & $\approx 99\%$ BERT\tnote{$\diamond$} \\
Q8BERT~\cite{zafrir2019q8bert} & ~ & 12 & DQ + QAT & - & & BERT$_{\mathrm{BASE}}$ & $\approx 99\%$ BERT\\
\midrule
ALBERT\tnote{$\mathsection$}~~\cite{lan2019albert} &Param. Sharing & 12 & $\mathcal{L}_{\textrm{\tiny MLM}}$ + $\mathcal{L}_{\textrm{\tiny SOP}}$ & $\times 5.6 \sim 0.3$ & 12 $\sim$ 235M && 89.4 (ensemble)\\
\midrule
DistilBERT~\cite{sanh2019distilbert} & \multirow{7}*{Distillation} & 6 & $\mathcal{L}_{\textrm{\tiny KD-CE}}$+Cos$_{\mathrm{KD}}$+ $\mathcal{L}_{\textrm{\tiny MLM}}$ & $\times 1.63$ & 66M & BERT$_{\mathrm{BASE}}$ & 77.0 (dev)\\
TinyBERT\tnote{$\mathsection$}~~\tnote{$\dagger$} ~\cite{jiao2019tinybert} && 4 & MSE$_{\mathrm{embed}}$+MSE$_{\mathrm{attn}}$+ MSE$_{\mathrm{hidn}}$+$\mathcal{L}_{\textrm{\tiny KD-CE}}$ & $\times 9.4$ & 14.5M & BERT$_{\mathrm{BASE}}$ & 76.5\\
BERT-PKD~\cite{sun-etal-2019-patient} & ~ & 3 $\sim$ 6 & $\mathcal{L}_{\textrm{\tiny KD-CE}}$+PT$_{\mathrm{KD}}$+ $\mathcal{L}_{\textrm{\tiny Task}}$ & $\times 3.73\sim 1.64$ & 45.7 $\sim$ 67 M & BERT$_{\mathrm{BASE}}$ & 76.0 $\sim$ 80.6\tnote{$\sharp$}\\
PD~\cite{turc2019well} & ~ & 6 & $\mathcal{L}_{\textrm{\tiny KD-CE}}$+$\mathcal{L}_{\textrm{\tiny Task}}$+ $\mathcal{L}_{\textrm{\tiny MLM}}$ & $\times 2.0$ & 67.5M & BERT$_{\mathrm{BASE}}$ & 81.2\tnote{$\sharp$}\\
MobileBERT\tnote{$\mathsection$}~\cite{sun2020mobilebert} & ~ & 24 & FMT+AT+PKT+ $\mathcal{L}_{\textrm{\tiny KD-CE}}$+$\mathcal{L}_{\textrm{\tiny MLM}}$ & $\times 4.0$ & 25.3M & BERT$_{\mathrm{LARGE}}$ & 79.7\\
MiniLM~\cite{wang2020minilm} & ~ & 6 & AT+AR & $\times 1.99$ & 66M & BERT$_{\mathrm{BASE}}$ & 81.0\tnote{$\flat$} \\
DualTrain\tnote{$\mathsection$}~~\tnote{$\dagger$}~\cite{zhao2019extreme} & ~ & 12 & Dual Projection+$\mathcal{L}_{\textrm{\tiny MLM}}$ & - & 1.8 $\sim$ 19.2M & BERT$_{\mathrm{BASE}}$ & 75.8 $\sim$ 81.9\tnote{$\natural$}\\
\midrule
BERT-of-Theseus~\cite{xu2020bert} & Module Replacing & 6 & $\mathcal{L}_{\textrm{\tiny Task}}$ & $\times1.94$& 66M& BERT$_{\mathrm{BASE}}$ & 78.6 \\
 \bottomrule
 \end{tabular}
\begin{tablenotes}
\item[1] The desing of this table is borrowed from \cite{xu2020bert,rogers2020primer}.
\item[$\ddagger$] The averaged score on 8 tasks (without WNLI) of GLUE benchmark (see Section \ref{sec:app-glue}). Here MNLI-m and MNLI-mm are regarded as two different tasks.
    `dev' indicates the result is on dev set. `ensemble' indicates the result is from the ensemble model.
 \item[$*$]  `$\mathcal{L}_{\textrm{\tiny MLM}}$ ', `$\mathcal{L}_{\textrm{\tiny NSP}}$', and `$\mathcal{L}_{\textrm{\tiny SOP}}$' indicate pre-training objective (see Section~\ref{sec:pre-training-tasks} and Table \ref{tab:loss-function}).`$\mathcal{L}_{\textrm{\tiny Task}}$' means task-specific loss.\\
     `HAWQ', `GWQ', `DQ', and `QAT' indicate Hessian AWare Quantization, Group-wise Quantization, Quantization-Aware Training, and Dynamically Quantized, respectively. `KD' means knowledge distillation. `FMT', `AT', and `PKT' mean Feature Map Transfer, Attention Transfer, and Progressive Knowledge Transfer, respectively. `AR' means Self-Attention value relation.
\item[$\mathsection$] The dimensionality of the hidden or embedding layers is reduced.
\item[$\dagger$] Use a smaller vocabulary.
\item[$\flat$] Generally, the F1 score is usually used as the main metric of the QQP task. But MiniLM reports the accuracy, which is incomparable to other works.
\item[$\diamond$] Result on MNLI and SST-2 only.
\item[$\sharp$] Result on the other tasks except for STS-B and CoLA.
 \item[$\natural$] Result on MRPC, MNLI, and SST-2 only.
\end{tablenotes}
 \end{threeparttable}
 \end{table*}

\subsubsection{Model Pruning}

Model pruning refers to removing part of neural network (e.g., weights, neurons, layers, channels, attention heads), thereby achieving the effects of reducing the model size and speeding up inference time.

\citet{gordon2020compressing} explored the timing of pruning (e.g., pruning during pre-training, after downstream fine-tuning) and the pruning regimes. \citet{michel2019sixteen} and \citet{voita-etal-2019-analyzing} tried to prune the entire self-attention heads in the transformer block.

\subsubsection{Quantization}
Quantization refers to the compression of higher precision parameters to lower precision. Works from \citet{shen2020q} and \citet{zafrir2019q8bert} solely focus on this area. Note that quantization often requires compatible hardware.

\subsubsection{Parameter Sharing}
Another well-known approach to reduce the number of parameters is parameter sharing, which is widely used in CNNs, RNNs, and Transformer~\cite{DBLP:conf/iclr/DehghaniGVUK19}.
ALBERT~\cite{lan2019albert} uses \textit{cross-layer parameter sharing} and \textit{factorized embedding parameterization}  to reduce the parameters of PTMs. Although the number of parameters is greatly reduced, the training and inference time of ALBERT are even longer than the standard BERT.

Generally, parameter sharing does not improve the computational efficiency at inference phase.

\subsubsection{Knowledge Distillation}

Knowledge distillation (KD)~\cite{hinton2015distilling} is a compression technique in which a small model called \textit{student model} is trained to reproduce the behaviors of a large model called \textit{teacher model}. Here the teacher model can be an ensemble of many models and usually well pre-trained. Different to model compression, distillation techniques learn a small student model from a fixed teacher model through some optimization objectives, while compression techniques aiming at searching a sparser architecture.

Generally, distillation mechanisms can be divided into three types: (1) distillation from soft target probabilities, (2) distillation from other knowledge, and (3) distillation to other structures:

(1) \textit{Distillation from soft target probabilities}.
\citet{bucilua2006model} showed that making the student approximate the teacher model can transfer knowledge from teacher to student. A common method is approximating the logits of the teacher model. DistilBERT~\cite{sanh2019distilbert} trained the student model with a distillation loss over the soft target probabilities of the teacher as:
\begin{align}\label{eq:distillation}
\mathcal{L}_{\textrm{\tiny KD-CE}}=\sum_{i}{t_i \cdot \mathrm{log}(s_i )},
\end{align}
where $t_i$ and $s_i$ are the probabilities estimated by the teacher model and the student, respectively.

Distillation from soft target probabilities can also be used in task-specific models, such as information retrieval~\cite{lu2020twinbert}, and sequence labeling~\cite{tsai-etal-2019-small}.

(2) \textit{Distillation from other knowledge}.
Distillation from soft target probabilities regards the teacher model as a black box and only focus on its outputs. Moreover, decomposing the teacher model and distilling more knowledge can bring improvement to the student model.

TinyBERT~\cite{jiao2019tinybert} performs layer-to-layer distillation with embedding outputs, hidden states, and self-attention distributions.    MobileBERT~\cite{sun2020mobilebert} also perform layer-to-layer distillation with soft target probabilities, hidden states, and self-attention distributions. MiniLM~\cite{wang2020minilm} distill self-attention distributions and self-attention value relation from teacher model.

Besides, other models distill knowledge through many approaches. \citet{sun-etal-2019-patient} introduced a ``\textit{patient}'' teacher-student mechanism, \citet{liu2019improving} exploited KD to improve a pre-trained multi-task deep neural network.

(3) \textit{Distillation to other structures}. Generally, the structure of the student model is the same as the teacher model, except for a smaller layer size and a smaller hidden size. However, not only decreasing parameters but also simplifying model structures from Transformer to RNN~\cite{tang2019distilling} or CNN~\cite{chia2019transformer} can reduce the computational complexity.

\subsubsection{Module Replacing}

Module replacing is an interesting and simple way to reduce the model size, which replaces the large modules of original PTMs with more compact substitutes.
\citet{xu2020bert} proposed Theseus Compression motivated by a famous thought experiment called ``Ship of Theseus'', which progressively substitutes modules from the source model with modules of fewer parameters.
Different from KD, Theseus Compression only requires one task-specific loss function.
The compressed model, BERT-of-Theseus, is $1.94\times$ faster while retaining more than 98\% performance of the source model.

\subsubsection{Early Exit}
Another efficient way to reduce the inference time is early exit, which allows the model to exit early at an off-ramp instead of passing through the entire model. The number of layers to be executed is conditioned on the input.

The idea of early exit is first applied in computer vision, such as BranchyNet~\cite{DBLP:conf/icpr/Teerapittayanon16} and Shallow-Deep Network~\cite{Kaya2019Shallow}. With the emergence of deep pre-trained language models, early exit is recently adopted to speedup Transformer-based models. As a prior work, Universal Transformer~\cite{DBLP:conf/iclr/DehghaniGVUK19} uses the Adaptive Computation Time (ACT) mechanism~\cite{Graves2016Adaptive} to achieve input-adaptive computation. \citet{Elbayad2020Depth} proposed Depth-adaptive transformer for machine translation, which learns to predict how many decoding layers are required for a particular sequence or token. Instead of learning how much computation is required, \citet{DBLP:conf/aaai/LiuMZCX21} proposed two estimation approaches based on Mutual Information (MI) and Reconstruction Loss respectively to directly allocate the appropriate computation to each sample.

More recently, DeeBERT~\cite{xin-etal-2020-deebert}, RightTool~\cite{DBLP:conf/acl/SchwartzSSDS20}, FastBERT~\cite{liu2020fastbert}, ELBERT~\cite{xie2021elbert}, PABEE~\cite{zhou2020bert} are proposed to reduce the computation of transformer encoder for natural language understanding tasks. Their methods usually contain two steps: (a) Training the injected off-ramps (aka internal classifiers), and (b) Designing an exiting strategy to decide whether or not to exit.

Typically, the training objective is a weighted sum of the cross-entropy losses at all off-ramps, i.e.
\begin{equation}
    \mathcal{L}_{\text{early-exit}} = \sum_{i=1}^{M} w_i\cdot \mathcal{L}_i,
\end{equation}
where $M$ is the number of off-ramps. FastBERT~\cite{liu2020fastbert} adopted the self-distillation loss that trains each off-ramp with the soft target generated by the final classifier. \citet{Liao2021Global} improved the objective by considering both the past and the future information. In particular, the off-ramps are trained to aggregate the hidden states of the past layers, and also approximate the hidden states of the future layers. Moreover, \citet{Sun2021Early} developed a novel training objective from the perspective of ensemble learning and mutual information, by which the off-ramps are trained as an ensemble. Their proposed objective not only optimizes the accuracy of each off-ramp but also the diversity of the off-ramps.

During inference, an exiting strategy is required to decide whether to exit early or continue to the next layer. DeeBERT~\cite{xin-etal-2020-deebert},  FastBERT~\cite{liu2020fastbert}, \citet{Liao2021Global} adopt the entropy of the prediction distribution as the exiting criterion. Similarly, RightTool~\cite{DBLP:conf/acl/SchwartzSSDS20} use the maximum softmax score to decide whether to exit. PABEE developed a patience-based strategy that allows a sample to exit when the prediction is unchanged for successive layers. Further, \citet{Sun2021Early} adopt a voting-based strategy to let all of the past off-ramps take a vote to decide whether or not to exit. Besides, \citet{li2021accelerating} proposed a window-based uncertainty as the exiting criterion to achieve token-level early exit (TokEE) for sequence labeling tasks.

\section{Adapting PTMs to Downstream Tasks}
\label{sec:adapt}

Although PTMs capture the general language knowledge from a large corpus, how effectively adapting their knowledge to the downstream task is still a key problem.

\subsection{Transfer Learning}

\begin{figure}[H]
  \centering
  \includegraphics[width=0.45\textwidth]{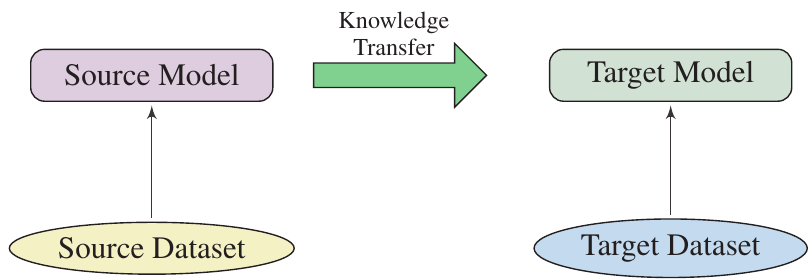}
  \caption{Transfer Learning}
  \label{fig:transfer-learning}
\end{figure}

Transfer learning~\cite{pan2009survey} is to adapt the knowledge from a source task (or domain) to a target task (or domain). Figure~\ref{fig:transfer-learning} gives an illustration of transfer learning.

There are many types of transfer learning in NLP, such as domain adaptation, cross-lingual learning, multi-task learning.
Adapting PTMs to downstream tasks is \textit{sequential transfer learning} task, in which tasks are learned sequentially and the target task has labeled data.

\subsection{How to Transfer?}

To transfer the knowledge of a PTM to the downstream NLP tasks, we need to consider the following issues:

\subsubsection{Choosing appropriate pre-training task, model architecture and corpus}

Different PTMs usually have different effects on the same downstream task, since these PTMs are trained with various pre-training tasks, model architecture, and corpora.

(1) Currently, the language model is the most popular pre-training task and can more efficiently solve a wide range of NLP problems ~\cite{radford2019language}.
However,
different pre-training tasks have their own bias and give different effects for different tasks.
For example, the NSP task \cite{devlin2019bert} makes PTM understand the relationship between two sentences. Thus, the PTM can benefit downstream tasks such as Question Answering (QA) and Natural Language Inference (NLI).

(2) The architecture of PTM is also important for the downstream task. For example, although BERT helps with most natural language understanding tasks, it is hard to generate language.

(3) The data distribution of the downstream task should be approximate to PTMs. Currently, there are a large number of off-the-shelf PTMs, which can just as conveniently be used for various domain-specific or language-specific downstream tasks.

Therefore, given a target task, it is always a good solution to choose the PTMs trained with appropriate pre-training task, architecture, and corpus.


\subsubsection{Choosing appropriate layers}

Given a pre-trained deep model, different layers should capture different kinds of information, such as POS tagging, parsing, long-term dependencies, semantic roles, coreference. For RNN-based models, \citet{belinkov-etal-2017-neural} and \citet{melamud-etal-2016-context2vec} showed that representations learned from different layers in a multi-layer LSTM encoder benefit different tasks (e.g., predicting POS tags and understanding word sense).
For transformer-based PTMs, \citet{DBLP:conf/acl/TenneyDP19} found BERT represents the steps of the traditional NLP pipeline: basic syntactic information appears
earlier in the network, while high-level semantic
information appears at higher layers.

Let $\bH^{(l)}( 1\leq l\leq L$) denotes the $l$-th layer representation of the pre-trained model with $L$ layers, and $g(\cdot)$ denote the task-specific model for the target task.

There are three ways to select the representation:

a) \textit{Embedding Only.}
One approach is to choose only the pre-trained static embeddings, while the rest of the model still needs to be trained from scratch for a new target task.

 They fail to capture higher-level information that might be even more useful. Word embeddings are only useful in capturing semantic meanings of words, but we also need to understand higher-level concepts like word sense.

b) \textit{Top Layer.} The most simple and effective way is to feed the representation at the top layer into the task-specific model $g(\bH^{(L)})$.

c) \textit{All Layers.} A more flexible way is to automatic choose the best layer in a soft version, like ELMo~\cite{peters2018elmo}:
\begin{align}
\br_t = \gamma \sum_{l=1}^{L} \alpha_l \bh_t^{(l)},
\end{align}
where $\alpha_l$ is the softmax-normalized weight for layer $l$ and $\gamma$ is a scalar to scale the vectors output by pre-trained model.
The mixup representation is fed into the task-specific model $g(\br_t)$.

\subsubsection{To tune or not to tune?}
Currently, there are two common ways of model transfer: feature extraction (where the pre-trained parameters are frozen), and fine-tuning (where the pre-trained parameters are unfrozen and fine-tuned).

In feature extraction way, the pre-trained models are regarded as off-the-shelf feature extractors.
Moreover, it is important to expose the internal layers as they typically encode the most transferable representations~\cite{peters2019tune}.

Although both these two ways can significantly benefit most of NLP tasks, feature extraction way requires more complex task-specific architecture. Therefore, the fine-tuning way is usually more general and convenient for many different downstream tasks than feature extraction way.

Table \ref{tab:layer-ft} gives some common combinations of adapting PTMs.

\begin{table}[H]
 \footnotesize\centering
 \begin{threeparttable}\caption{Some common combinations of adapting PTMs.}\label{tab:layer-ft}
 \doublerulesep 0.1pt
 \tabcolsep 5pt 
 \begin{tabular}{ccP{0.5\linewidth}}
 \toprule
  Where   &  FT/FE?\tnote{$\dagger$} & PTMs \\
 \midrule
  Embedding Only & FT/FE & Word2vec \cite{mikolov2013word2vec}, GloVe \cite{DBLP:conf/emnlp/PenningtonSM14}\\
  Top Layer & FT & BERT~\cite{devlin2019bert}, RoBERTa~\cite{liu2019roberta}\\
  Top Layer & FE & BERT\tnote{$\mathsection$}~~\cite{zhong-etal-2019-searching,Zhu2020Incorporating}\\
  All Layers & FE & ELMo~\cite{peters2018elmo} \\
 \bottomrule
 \end{tabular}
\begin{tablenotes}
\item[$\dagger$] FT and FE mean Fine-tuning and Feature Extraction respectively.
\item[$\mathsection$] BERT used as feature extractor.
\end{tablenotes}
 \end{threeparttable}
 \end{table}


\subsection{Fine-Tuning Strategies}

With the increase of the depth of PTMs, the representation captured by them makes the downstream task easier. Therefore, the task-specific layer of the whole model is simple.
Since ULMFit and BERT, fine-tuning has become the main adaption method of PTMs.
However, the process of fine-tuning is often brittle: even with the same hyper-parameter values, distinct random seeds can lead to substantially different results~\cite{dodge2020fine}.

Besides standard fine-tuning, there are also some useful fine-tuning strategies.

\paragraph{Two-stage fine-tuning}

An alternative solution is \textit{two-stage transfer}, which introduces an intermediate stage between pre-training and fine-tuning.
In the first stage, the PTM is transferred into a model fine-tuned by an intermediate task or corpus. In the second stage, the transferred model is fine-tuned to the target task.
\citet{sun2019fine} showed that the ``further pre-training'' on the related-domain corpus can further improve the ability of BERT and achieved state-of-the-art performance on eight widely-studied text classification datasets.
\citet{phang2018sentence} and \citet{garg2019tanda} introduced the intermediate supervised task related to the target task, which brings a large improvement for BERT, GPT, and ELMo. \citet{li2019story} also used a two-stage transfer for the story ending prediction.
The proposed TransBERT (transferable BERT) can transfer not only general language knowledge from large-scale unlabeled data but also specific kinds of knowledge from various semantically related supervised tasks.


\paragraph{Multi-task fine-tuning}
\citet{liu2019multi} fine-tuned BERT under the multi-task learning framework, which demonstrates that multi-task learning and pre-training are complementary technologies.

\paragraph{Fine-tuning with extra adaptation modules}

The main drawback of fine-tuning is its parameter inefficiency: every downstream task has its own fine-tuned parameters.
Therefore, a better solution is to inject some fine-tunable adaptation modules into PTMs while the original parameters are fixed.

\citet{stickland2019bert} equipped a single share BERT model with small additional task-specific adaptation modules, projected attention layers (PALs). The shared BERT with the PALs matches separately fine-tuned models on the GLUE benchmark with roughly 7 times fewer parameters.
Similarly, \citet{houlsby2019parameter} modified the architecture of pre-trained BERT by adding adapter modules. Adapter modules yield a compact and extensible model; they add only a few trainable parameters per task, and new tasks can be added without revisiting previous ones. The parameters of the original network remain fixed, yielding a high degree of parameter sharing.

\paragraph{Others}

Motivated by the success of widely-used ensemble models, \citet{xu2020improving} improved the fine-tuning of BERT with two effective mechanisms: \textit{self-ensemble} and \textit{self-distillation}, which can improve the performance of BERT on downstream tasks without leveraging external resource or significantly decreasing the training efficiency.
They integrated ensemble and distillation within a single training process. The teacher model is an ensemble model by parameter-averaging several student models in previous time steps.

Instead of fine-tuning all the layers simultaneously, \textit{gradual unfreezing}~\cite{DBLP:conf/acl/RuderH18} is also an effective method that gradually unfreezes layers of PTMs starting from the top layer.
\citet{chronopoulou2019embarrassingly} proposed a simpler unfreezing method, \textit{sequential unfreezing}, which first fine-tunes only the randomly-initialized task-specific layers, and then unfreezes the hidden layers of PTM, and finally unfreezes the embedding layer.

\citet{li-eisner-2019-specializing} compressed ELMo embeddings using variational information bottleneck while keeping only the information that helps the target task.

Generally, the above works show that the utility of PTMs can be further stimulated by better fine-tuning strategies.



\subsubsection{Prompt-based Tuning}
Narrowing the gap between pre-training and fine-tuning can further boost the performance of PTMs on downstream tasks. An alternative approach is reformulating the downstream tasks into a MLM task by designing appropriate prompts. Prompt-based methods have shown great power in few-shot setting~\cite{Scao2021How,Schick2021Exploting,Schick2021PET,Gao2020LMBFF}, zero-shot setting~\cite{petroni2019language,Jiang2020How}, and even fully-supervised setting~\cite{Li2021Prefix,Liu2021PTuning}. Current prompt-based methods can be categorized as two branches according to the prompt is whether discrete or continuous.

\paragraph{Discrete prompts} Discrete prompt is a sequence of words to be inserted into the input text, which helps the PTM to better model the downstream task. 
\citet{sun2019utilizing} constructed an auxiliary sentence by transforming aspect-based sentiment analysis (ABSA) task to a sentence pair classification task, but its model parameters still need to be fine-tuned. 
GPT-3~\cite{Brown2020GPT3} proposed the in-context learning that concatenates the original input with the task description and a few examples. By this, GPT-3 can achieve competitive performance without tuning the parameters. Besides, \citet{petroni2019language} found that with proper manual prompt, BERT can perform well on entity prediction task (LAMA) without training. In addition to LAMA, ~\citet{Schick2021Exploting,Schick2021PET} proposed PET that designed discrete prompts for various text classification and entailment tasks. However, the manually designed prompts can be sub-optimal, as a result, many methods are developed to automate the generation of prompts. LPAQA~\cite{Jiang2020How} uses two methods, i.e., mining-based generation and paraphrasing-based generation, to find the optimal patterns that express particular relations. AutoPrompt~\cite{Shin2020AutoPrompt} finds the optimal prompt with gradient-guided search. LM-BFF~\cite{Gao2020LMBFF} employs T5~\cite{raffel2019t5} to automatically generate prompts.

\paragraph{Continuous prompts} Instead of finding the optimal concrete prompt, another alternative is to directly optimize the prompt in continuous space, i.e. the prompt vectors are not necessarily word type embeddings of the PTM. The optimized continuous prompt is concatenated with word type embeddings, which is then fed into the PTM. \citet{Qin2021Learning} and \citet{Zhong2021Factual} found that the optimized continuous prompt can outperform concrete prompts (including manual~\cite{petroni2019language}, mined (LPAQA~\cite{Jiang2020How}), and gradient-searched (AutoPrompt~\cite{Shin2020AutoPrompt}) prompts) on relational tasks. WARP~\cite{Hambardzumyan2021WARP} inserts trainable continuous prompt tokens before, between, and after the input sequence while keeping the parameters of the PTM fixed, resulting in considerable performance on GLUE benchmark. Prefix-Tuning~\cite{Li2021Prefix} inserts continuous prompt as prefix of the input of GPT-2 for table-to-text generation and BART for summarization. Prefix-Tuning, as a parameter-efficient tuning technique, achieved comparable performance in fully-supervised setting and outperformed model fine-tuning in few-shot setting. Further, P-Tuning~\cite{Liu2021PTuning} showed that, with continuous prompt, GPT can also achieve comparable or even better performance to similar-sized BERT on natural language understanding (NLU) tasks. Very recently, \citet{Lester2021Power} showed that prompt tuning becomes more competitive with scale. When the PTM exceeds billions of parameters, the gap between model fine-tuning and prompt tuning can be closed, which makes the prompt-based tuning a very promising method for efficient serving of large-scale PTMs.

\section{Resources of PTMs}
\label{sec:resources}

There are many related resources for PTMs available online. Table~\ref{tab:ptms-resources} provides some popular repositories, including third-party implementations, paper lists, visualization tools, and other related resources of PTMs.

\begin{table*}[ht]
 \scriptsize\centering
 \begin{threeparttable}\caption{Resources of PTMs}
 \label{tab:ptms-resources}
 \begin{tabular}{l@{\hskip 1em}P{30em}@{\hskip 1em}P{27em}}
 \toprule
  Resource &  Description & URL \\
 \midrule
 \multicolumn{3}{c}{Open-Source Implementations~\tnote{$\mathsection$}}\\
 \midrule
 word2vec &  CBOW,Skip-Gram & https://github.com/tmikolov/word2vec\\
 GloVe & Pre-trained word vectors & https://nlp.stanford.edu/projects/glove\\
 FastText &  Pre-trained word vectors & https://github.com/facebookresearch/fastText\\
 \midrule
  Transformers & Framework: PyTorch\&TF, \ \ PTMs: BERT, GPT-2, RoBERTa, XLNet, etc. & https://github.com/huggingface/transformers\\
  Fairseq & Framework: PyTorch, \ \ PTMs:English LM, German LM, RoBERTa, etc. & https://github.com/pytorch/fairseq \\
  Flair & Framework: PyTorch, \ \ PTMs:BERT, ELMo, GPT, RoBERTa, XLNet, etc. &https://github.com/flairNLP/flair \\
  AllenNLP~\cite{Gardner2017AllenNLP} & Framework: PyTorch, \ \ PTMs: ELMo, BERT, GPT-2, etc. & https://github.com/allenai/allennlp\\
  fastNLP & Framework: PyTorch, \ \ PTMs: RoBERTa, GPT, etc. & https://github.com/fastnlp/fastNLP \\
  UniLMs  &Framework: PyTorch, \ \  PTMs: UniLM v1\&v2, MiniLM, LayoutLM, etc. & https://github.com/microsoft/unilm\\
 \midrule
  Chinese-BERT~\cite{cui2019pretraining} & Framework: PyTorch\&TF, \ \ PTMs: BERT, RoBERTa, etc. (for Chinese) & https://github.com/ymcui/Chinese-BERT-wwm \\
  BERT~\cite{devlin2019bert} & Framework: TF, \ \ PTMs: BERT, BERT-wwm & https://github.com/google-research/bert \\
  RoBERTa~\cite{liu2019roberta} & Framework: PyTorch & https://github.com/pytorch/fairseq/tree/master/examples/roberta \\
  XLNet~\cite{yang2019xlnet} &Framework: TF& https://github.com/zihangdai/xlnet/ \\
  ALBERT~\cite{lan2019albert} & Framework: TF & https://github.com/google-research/ALBERT \\
  T5~\cite{raffel2019t5} & Framework: TF & https://github.com/google-research/text-to-text-transfer-transformer \\
  ERNIE(Baidu)~\cite{sun2019ernie,sun2019ernie2} & Framework: PaddlePaddle  & https://github.com/PaddlePaddle/ERNIE \\
  CTRL~\cite{keskar2019ctrl} & Conditional Transformer Language Model for Controllable Generation. & https://github.com/salesforce/ctrl\\
\midrule
BertViz~\cite{vig2019transformervis} & Visualization Tool & https://github.com/jessevig/bertviz \\
exBERT~\cite{hoover2019exbert} & Visualization Tool&
https://github.com/bhoov/exbert\\
\midrule
TextBrewer~\cite{textbrewer} & PyTorch-based toolkit for distillation of NLP models. & https://github.com/airaria/TextBrewer\\
DeepPavlov & Conversational AI Library. PTMs for the Russian, Polish, Bulgarian, Czech, and informal English. & https://github.com/deepmipt/DeepPavlov\\
   \toprule
 \multicolumn{3}{c}{Corpora}\\
  \midrule
OpenWebText & Open clone of OpenAI's unreleased WebText dataset. & https://github.com/jcpeterson/openwebtext\\
Common Crawl& A very large collection of text. & http://commoncrawl.org/\\
WikiEn & English Wikipedia dumps. &https://dumps.wikimedia.org/enwiki/\\
  \toprule
 \multicolumn{3}{c}{Other Resources}\\
  \midrule
 Paper List& & https://github.com/thunlp/PLMpapers\\
 Paper List& & https://github.com/tomohideshibata/BERT-related-papers\\
 Paper List& & https://github.com/cedrickchee/awesome-bert-nlp\\
 Bert Lang Street& A collection of BERT models with reported performances on different datasets, tasks and languages. & https://bertlang.unibocconi.it/\\
 \bottomrule
 \end{tabular}
 \begin{tablenotes}
\item[$\mathsection$] Most papers for PTMs release their links of official version. Here we list some popular third-party and official implementations.
\end{tablenotes}
 \end{threeparttable}
\end{table*}

Besides, there are some other good survey papers on PTMs for NLP~\cite{wang2020static,liu2020survey,rogers2020primer}.


%
%
%

\section{Applications}
\label{sec:app}

In this section, we summarize some applications of PTMs in several classic NLP tasks.

\subsection{General Evaluation Benchmark}
\label{sec:app-glue}

There is an essential issue for the NLP community that how can we evaluate PTMs in a comparable metric. Thus, large-scale-benchmark is necessary.

The General Language Understanding Evaluation (GLUE) benchmark~\cite{DBLP:conf/iclr/WangSMHLB19} is a collection of nine natural language understanding tasks, including single-sentence classification tasks (CoLA and SST-2), pairwise text classification tasks (MNLI, RTE, WNLI, QQP, and MRPC), text similarity task (STS-B), and relevant ranking task (QNLI). GLUE benchmark is well-designed for evaluating the robustness as well as generalization of models. GLUE does not provide the labels for the test set but set up an evaluation server.

However, motivated by the fact that the progress in recent years has eroded headroom on the GLUE benchmark dramatically, a new benchmark called SuperGLUE~\cite{DBLP:conf/nips/WangPNSMHLB19} was presented. Compared to GLUE, SuperGLUE has more challenging tasks and more diverse task formats (e.g., coreference resolution and question answering).

State-of-the-art PTMs are listed in the corresponding leaderboard\footnote{https://gluebenchmark.com/} ~\footnote{https://super.gluebenchmark.com/}.

\subsection{Question Answering}

Question answering (QA), or a narrower concept machine reading comprehension (MRC), is an important application in the NLP community. From easy to hard, there are three types of QA tasks: single-round extractive QA (SQuAD)~\cite{rajpurkar2016squad}, multi-round generative QA (CoQA)~\cite{DBLP:journals/tacl/ReddyCM19}, and multi-hop QA (HotpotQA)~\cite{DBLP:conf/emnlp/Yang0ZBCSM18}.

BERT creatively transforms the extractive QA task to the spans prediction task that predicts the starting span as well as the ending span of the answer~\cite{devlin2019bert}. After that, PTM as an encoder for predicting spans has become a competitive baseline. For extractive QA, \citet{zhang2020retrospective} proposed a retrospective reader architecture and initialize the encoder with PTM (e.g., ALBERT). For multi-round generative QA, \citet{ju2019technical} proposed a ``PTM+Adversarial Training+Rationale Tagging+Knowledge Distillation" model. For multi-hop QA, \citet{tu2020select} proposed an interpretable ``Select, Answer, and Explain" (SAE) system that PTM acts as the encoder in the selection module.

Generally, encoder parameters in the proposed QA model are initialized through a PTM, and other parameters are randomly initialized. State-of-the-art models are listed in the corresponding leaderboard. \footnote{https://rajpurkar.github.io/SQuAD-explorer/}
~\footnote{https://stanfordnlp.github.io/coqa/}
~\footnote{https://hotpotqa.github.io/}

\subsection{Sentiment Analysis}

BERT outperforms previous state-of-the-art models by simply fine-tuning on SST-2, which is a widely used dataset for sentiment analysis (SA) \cite{devlin2019bert}. \citet{bataa2019investigation} utilized BERT with transfer learning techniques and achieve new state-of-the-art in Japanese SA.

Despite their success in simple sentiment classification, directly applying BERT to aspect-based sentiment analysis (ABSA), which is a fine-grained SA task, shows less significant improvement~\cite{sun2019utilizing}. To better leverage the powerful representation of BERT, \citet{sun2019utilizing} constructed an auxiliary sentence by transforming ABSA from a single sentence classification task to a sentence pair classification task. \citet{xu2019post} proposed post-training to adapt BERT from its source domain and tasks to the ABSA domain and tasks. Furthermore, \citet{rietzler2019adapt} extended the work of \cite{xu2019post} by analyzing the behavior of cross-domain post-training with ABSA performance. \citet{karimi2020adversarial} showed that the performance of post-trained BERT could be further improved via adversarial training. \citet{song2020utilizing} added an additional pooling module, which can be implemented as either LSTM or attention mechanism, to leverage BERT intermediate layers for ABSA. In addition, \citet{li2019exploiting} jointly learned aspect detection and sentiment classification towards end-to-end ABSA. SentiLR~\cite{ke2019sentilr} acquires part-of-speech tag and prior sentiment
polarity from SentiWordNet and adopts \textit{Label-Aware MLM} to utilize the
introduced linguistic knowledge to capture the
relationship between sentence-level sentiment
labels and word-level sentiment shifts. SentiLR achieves state-of-the-art performance on several sentence- and aspect-level sentiment classification tasks.

For sentiment transfer, \citet{wu2019mask} proposed ``Mask and Infill" based on BERT. In the mask step, the model disentangles sentiment from content by masking sentiment tokens. In the infill step, it uses BERT along with a target sentiment embedding to infill the masked positions.

%
%

\subsection{Named Entity Recognition}

Named Entity Recognition (NER)  in information extraction and plays an important role in many NLP downstream tasks. In deep learning, most of NER methods are in the sequence-labeling framework. The entity information in a sentence will be transformed into the sequence of labels, and one label corresponds to one word. The model is used to predict the label of each word. Since ELMo and BERT have shown their power in NLP, there is much work about pre-trained models for NER.

\citet{akbik2018contextual} used a pre-trained character-level language model to produce word-level embedding for NER.
TagLM \citep{DBLP:conf/acl/PetersABP17} and ELMo \citep{peters2018elmo} use a pre-trained language model's last layer output and weighted-sum of each layer output as a part of word embedding.
\citet{DBLP:conf/emnlp/LiuRSG0018} used layer-wise pruning and dense connection to speed up ELMo's inference on NER.
\citet{devlin2019bert} used the first BPE's BERT representation to predict each word's label without CRF. \citet{DBLP:journals/corr/abs-1906-01502} realized zero-shot NER through multilingual BERT. \citet{tsai-etal-2019-small} leveraged knowledge distillation to run a small BERT for NER on a single CPU. Besides, BERT is also used on domain-specific NER, such as biomedicine~\cite{hakala-pyysalo-2019-biomedical,lee2019biobert}, etc.

\subsection{Machine Translation}

Machine Translation (MT) is an important task in the NLP community, which has attracted many researchers. Almost all of Neural Machine Translation (NMT) models share the encoder-decoder framework, which first encodes input tokens to hidden representations
by the encoder and then decodes output tokens in the target language from the decoder.
\citet{ramachandran2017unsupervised} found the encoder-decoder models can be significantly improved by initializing both encoder and decoder with pre-trained weights of two language models. \citet{DBLP:conf/naacl/EdunovBA19} used ELMo to set the word embedding layer in the NMT  model. This work shows performance improvements on English-Turkish and English-German NMT model by using a pre-trained language model for source word embedding initialization.

Given the superb performance of BERT on other NLP tasks, it is natural to investigate how to incorporate BERT into NMT models.
\citet{DBLP:conf/nips/ConneauL19} tried to initialize the entire encoder and decoder by a  multilingual pre-trained BERT model and showed a significant improvement could be achieved on unsupervised  MT and English-Romanian supervised MT. Similarly, \citet{clinchant-etal-2019-use} devised a series of different experiments for examining the best strategy to utilize BERT on the encoder part of NMT models. They achieved some improvement by using BERT as an initialization of the encoder. Also, they found that these models can get better performance on the out-of-domain dataset. \citet{imamura-sumita-2019-recycling} proposed a two stages BERT fine-tuning method for NMT. At the first stage, the encoder is initialized by a pre-trained BERT model, and they only train the decoder on the training set. At the second stage, the whole NMT model is jointly fine-tuned on the training set. By experiment, they show this approach can surpass the one stage fine-tuning method, which directly fine-tunes the whole model. Apart from that, \citet{Zhu2020Incorporating} suggested using pre-trained BERT as an extra memory to facilitate NMT models. Concretely, they first encode the input tokens by a pre-trained BERT and use the output of the last layer as extra memory. Then, the NMT model can access the memory via an extra attention module in each layer of both encoder and decoder. And they show a noticeable improvement in supervised, semi-supervised, and unsupervised MT.

Instead of only pre-training the encoder, MASS (Masked Sequence-to-Sequence Pre-Training)~\cite{DBLP:conf/icml/SongTQLL19} utilizes Seq2Seq MLM to pre-train the encoder and decoder jointly. In the experiment, this approach can surpass the BERT-style pre-training proposed by \citet{DBLP:conf/nips/ConneauL19} both on unsupervised MT and English-Romanian supervised MT.
Different from MASS, mBART~\cite{liu2020multilingual}, a multilingual extension of BART~\cite{lewis2019bart}, pre-trains the encoder and decoder jointly with Seq2Seq denoising auto-encoder (DAE) task on large-scale monolingual corpora across 25 languages. Experiments demonstrated that mBART could significantly improve both supervised and unsupervised machine translation at
both the sentence level and document level.

\subsection{Summarization}

Summarization, aiming at producing a shorter text which preserves the most meaning of a longer text, has attracted the attention of the NLP community in recent years. The task has been improved significantly since the widespread use of PTM. \citet{zhong-etal-2019-searching} introduced transferable knowledge (e.g., BERT) for summarization and surpassed previous models. \citet{zhang-etal-2019-hibert} tries to pre-trained a document-level model that predicts sentences instead of words, and then apply it on downstream tasks such as summarization. More elaborately, \citet{zhang2019pegasus} designed a Gap Sentence Generation (GSG) task for pre-training, whose objective involves generating summary-like text from the input. Furthermore, \citet{liu2019text} proposed BERTSUM. BERTSUM included a novel document-level encoder, and a general framework for both extractive summarization and abstractive summarization. In the encoder frame, BERTSUM extends BERT by inserting multiple \texttt{[CLS]} tokens to learn the sentence representations. For extractive summarization, BERTSUM stacks several inter-sentence Transformer layers. For abstractive summarization, BERTSUM proposes a two-staged fine-tuning approach using a new fine-tuning schedule. \citet{zhong2020extractive} proposed a novel summary-level framework MATCHSUM and conceptualized extractive summarization as a semantic text
matching problem. They proposed a Siamese-BERT architecture to compute the similarity between the source document and the candidate summary and achieved a state-of-the-art result on CNN/DailyMail (44.41 in ROUGE-1) by only using the base version of BERT.

\subsection{Adversarial Attacks and Defenses}
\label{sec:attacks}

The deep neural models are vulnerable to adversarial examples that can mislead a model to produce a specific wrong prediction with imperceptible perturbations from the original input.
In CV, adversarial attacks and defenses have been widely studied. However, it is still challenging for text due to the discrete nature of languages.
Generating of adversarial samples for text needs to possess such qualities: (1) imperceptible to human judges yet misleading to neural models; (2) fluent in grammar and semantically consistent with original inputs.
\citet{jin2019bert} successfully attacked the fine-tuned BERT on text
classification and textual entailment with adversarial examples.
\citet{wallace2019universal} defined universal adversarial triggers that can induce a model to produce a specific-purpose prediction when concatenated to any input. Some triggers can even cause the GPT-2 model to generate racist text.
\citet{sun2020adv} showed BERT is not robust on misspellings.

PTMs also have great potential to generate adversarial samples.
\citet{li2020bertattack} proposed \textbf{BERT-Attack}, a BERT-based high-quality and effective attacker. They turned BERT against another fine-tuned BERT on downstream tasks and successfully misguided the target model to predict incorrectly, outperforming state-of-the-art attack strategies in both success rate and perturb percentage, while the generated adversarial samples are fluent and semantically preserved.


Besides, adversarial defenses for PTMs are also promising, which improve the robustness of PTMs and make them immune against adversarial attack.

Adversarial training aims to improve the generalization by minimizes the maximal risk for label-preserving perturbations in embedding space.
Recent work~\cite{zhu2020freelb,liu2020adversarial} showed that adversarial pre-training or fine-tuning can improve both generalization and
robustness of PTMs for NLP.

\section{Future Directions}
\label{sec:future}

Though PTMs have proven their power for various NLP tasks, challenges still exist due to the complexity of language. In this section, we suggest five future directions of PTMs.

\paragraph{(1) Upper Bound of PTMs}
Currently, PTMs have not yet reached its upper bound. Most of the current PTMs can be further improved by more training steps and larger corpora.

The state of the art in NLP can be further advanced by increasing the depth of models, such as Megatron-LM~\cite{shoeybi2019megatron} (8.3 billion parameters, 72 Transformer layers with a hidden size of 3072 and 32 attention heads) and Turing-NLG\footnote{https://www.microsoft.com/en-us/research/blog/turing-nlg-a-17-billion-parameter-language-model-by-microsoft/} (17 billion parameters, 78 Transformer layers with a hidden size of 4256 and 28 attention heads).

The general-purpose PTMs are always our pursuits for learning the intrinsic universal knowledge of languages (even world knowledge). However, such PTMs usually need deeper architecture, larger corpus, and challenging pre-training tasks, which further result in higher training costs.
However, training huge models is also a challenging problem, which needs more sophisticated and efficient training techniques such as distributed training, mixed precision, gradient accumulation, etc.
Therefore, a more practical direction is to design more efficient model architecture, self-supervised pre-training tasks, optimizers, and training skills using existing hardware and software. ELECTRA~\cite{clark2020electra} is a good solution towards this direction.

\paragraph{(2) Architecture of PTMs}

The Transformer has been proved to be an effective architecture for pre-training. However, the main limitation of the Transformer is its computation complexity, which is quadratic to the input length. Limited by the memory of GPUs, most of current PTMs cannot deal with the sequence longer than 512 tokens.
Breaking this limit needs to improve the architecture of the Transformer. Although many works~\cite{lin2021survey} tried to improve the efficiency of Transformer, there remains much room for improvement.

Besides, searching for more efficient alternative non-Transformer architecture for PTMs is important to capture longer-range contextual information.
The design of deep architecture is challenging, and we may seek help from some automatic methods, such as neural architecture search (NAS)~\cite{zoph2016neural}.

\paragraph{(3) Task-oriented Pre-training and Model Compression}

In practice, different downstream tasks require the different abilities of PTMs. The discrepancy between PTMs and downstream tasks usually lies in two aspects: model architecture and data distribution.
A larger discrepancy may result in that the benefit of PTMs may be insignificant.
For example, text generation usually needs a specific task to pre-train both the encoder and decoder, while text matching needs pre-training tasks designed for sentence pairs.

Besides, although larger PTMs can usually lead to better performance, a practical problem is how to leverage these huge PTMs on special scenarios, such as low-capacity devices and low-latency applications.
Therefore, we can carefully design the specific model architecture and pre-training tasks for downstream tasks or extract partial task-specific knowledge from existing PTMs.

Instead of training task-oriented PTMs from scratch, we can teach them with existing general-purpose PTMs by using techniques such as model compression (see Section \ref{sec:model-compression}). Although model compression is widely studied for CNNs in CV~\cite{cheng2017survey}, compression for PTMs for NLP is just beginning. The fully-connected structure of the Transformer also makes model compression more challenging.



\paragraph{(4) Knowledge Transfer Beyond Fine-tuning}

Currently, fine-tuning is the dominant method to transfer PTMs' knowledge to downstream tasks, but one deficiency is its parameter inefficiency: every downstream task has its own fine-tuned parameters. An improved solution is to fix the original parameters of PTMs and by adding small fine-tunable adaption modules for specific task~\cite{stickland2019bert,houlsby2019parameter}.
Thus, we can use a shared PTM to serve multiple downstream tasks.
Indeed, mining knowledge from PTMs can be more flexible, such as feature extraction, knowledge distillation~\cite{textbrewer}, data augmentation~\cite{wu2019conditional,kumar2020data}, using PTMs as external knowledge~\cite{petroni2019language}.
More efficient methods are expected.



\paragraph{(5) Interpretability and Reliability of PTMs}

Although PTMs reach impressive performance, their deep non-linear architecture makes the procedure of decision-making highly non-transparent.

Recently, explainable artificial intelligence (XAI)~\cite{arrieta2020explainable} has become a hotspot in the general AI community. Unlike CNNs for images, interpreting PTMs is harder due to the complexities of both the Transformer-like architecture and language.
Extensive efforts (see Section~\ref{sec:model-analysis}) have been made to analyze the linguistic and world knowledge included in PTMs, which help us understand these PMTs with some degree of transparency.
However, much work on model analysis depends
on the attention mechanism, and the effectiveness of attention for interpretability is still controversial ~\cite{jain2019attention,serrano2019attention}.

Besides, PTMs are also  vulnerable to adversarial attacks (see Section \ref{sec:attacks}). The reliability of PTMs is also becoming an issue of great concern with the extensive use of PTMs in production systems.
The studies of adversarial attacks against PTMs help us understand their capabilities by fully exposing their vulnerabilities. Adversarial defenses for PTMs are also promising, which improve the robustness of PTMs and make them immune against adversarial attack.

Overall, as key components in many NLP applications, the interpretability and reliability  of PTMs remain to be explored further in many respects, which helps us
understand how PTMs work and provides a guide for better usage and further improvement.


\section{Conclusion}
\label{sec:conclusion}

In this survey, we conduct a comprehensive overview of PTMs for NLP, including background knowledge, model architecture, pre-training tasks, various extensions, adaption approaches, related resources, and applications. Based on current PTMs, we propose a new taxonomy of PTMs from four different perspectives. We also suggest several possible future research directions for PTMs.

\section*{Acknowledgements}
We thank Zhiyuan Liu, Wanxiang Che, Minlie Huang, Danqing Wang and Luyao Huang for their valuable feedback on this manuscript.
This work was supported by the National Natural Science Foundation of China (No. 61751201 and 61672162), Shanghai Municipal Science and Technology Major Project (No. 2018SHZDZX01) and ZJLab.

%


\small
\setlength{\bibsep}{0.1cm}
\bibliography{nlp}
\bibliographystyle{unsrtnat}

\end{multicols}

\end{document}